%% file: main.tex
\documentclass{article}
\input{macros.tex}

\title{Training Large Scale Polynomial CNNs for E2E Inference over Homomorphic Encryption}
\author{Moran Baruch  $^{1}$ $^{2}$
\quad Nir Drucker  $^{1}$
\quad Gilad Ezov$^1$ 
\quad Yoav Goldberg$^{2}$ $^{3}$ 
\quad Eyal Kushnir$^{1}$ \\ 
\quad \textbf{Jenny Lerner}$^{1}$ 
\quad \textbf{Omri Soceanu}$^1$ 
\quad \textbf{Itamar Zimerman}$^1$ \\
$^1${IBM Research} \quad $^2${Bar-Ilan University} \quad $^3${AI2}\\
}
\begin{document}

\maketitle

\begin{abstract}
Training large-scale CNNs that during inference can be run under Homomorphic Encryption (HE) is challenging due to the need to use only polynomial operations. This limits HE-based solutions adoption.
We address this challenge and pioneer in providing a novel training method for large polynomial CNNs such as ResNet-152 and ConvNeXt models, and achieve promising accuracy on encrypted samples on large-scale dataset such as ImageNet. Additionally, we provide optimization insights regarding activation functions and skip-connection latency impacts, enhancing HE-based evaluation efficiency.
Finally, to demonstrate the robustness of our method, we provide a polynomial adaptation of the CLIP model for secure zero-shot prediction, unlocking unprecedented capabilities at the intersection of HE and transfer learning.

\end{abstract}

\section{Introduction} \label{sec:intro}
We are interested in the problem of training \glspl{CNN} in a way that allows inference on encrypted data, without the owner of the model being exposed to either the inputs or the outputs. This is achievable through \gls{HE}, see e.g., \cite{chet_compiler, helayers, baruch2021fighting}. Most modern HE schemes, however, limit network operations to polynomials, creating training and inference challenges.

Training large-scale polynomial \glspl{CNN} is a challenging task that often fails to achieve the same performance as the original network. Thus,
previous studies have only achieved promising results on shallow networks \cite{CryptoNets2016, baruch2021fighting}
%or very limited, toy-like datasets such as MNIST or CIFAR-10 \cite{lee2021precise, pmlr-v162-lee22e}, 
and their methods have not scaled up for larger networks or large datasets such as ImageNet \cite{deng2009imagenet}.

The hardness of training polynomial networks is well established, as we explain in Sect. \ref{sec:background}. 
Previous studies in HE (e.g., \cite{lee2021precise, pmlr-v162-lee22e, hyphen}) focused on modifying pre-trained networks by substituting non-polynomial ReLU activations for polynomial approximations,
however, %
when applied naively to deep networks, it can cause explosions or imprecise results. In this study, we observe that one of the factors for those explosions is the input range of the activation, which dominates the approximation error (Fig. \ref{fig:MiniMaxErrorperRange}). Hence the approximation requires extremely high-degree polynomials, leading to computational inefficiencies or instability. To address this, we develop a novel training method that handles the input range during the fine-tuning process, which enables approximating activations using low-degree polynomials. This method allows for the first time the training of \textbf{HE-friendly CNNs} on large-scale networks like ResNet \cite{resnet} and ConvNeXt~\cite{convnext1}, over large datasets such as ImageNet.

Another challenge for running encrypted inference is reducing the inference latency costs, which are significantly influenced by two key factors: the \textit{multiplication depth} of high-degree polynomials and the HE \textit{chain-index} mismatch resulting from skip-connections, as will be identified in Obs. \ref{obs:sc}.
% The latency costs of inference over encrypted data are significantly influenced by two key factors: the \textit{multiplication depth} of high-degree polynomials and the HE \textit{chain-index} mismatch resulting from skip connections.

To this end, we propose new design and training techniques for polynomial CNNs that translate to latency acceleration of the inference process of polynomial \glspl{CNN} under \gls{HE}.
Specifically, in Sect. \ref{sec:sc} we provide a solution to efficiently handle skip-connections under HE, through chain index-aware design, resulting in a substantial reduction in inference time. For instance, when employing the HElayers SDK, a notable speedup factor of 2.5 is achieved. %
% and a solution to reduce the number of skip connections that can significantly reduce inference time in Section \ref{sec:sc}.%
Additionally, in Sect. \ref{sec:backbones}, the paper provides an analysis of selecting an appropriate backbone to minimize the computational resources needed when using \gls{HE}.  %

% Another challenge in utilizing the polynomial NN under HE during inference is related to the latency costs, namely the multiplication depth that affected by the polynomial degree, and the skip connections, that due to chain index mismatch increase the latency costs (see Fig. \ref{fig:profiling}). Hence in Sect. \ref{sec:sc} and \ref{sec:backbones} will also demonstrate acceleration methods to achieve higher latency efficiency during secure inference.

% % Another limitation of HE is a relatively long inference time and expensive computation under HE, as will be detailed below.

% The first aspect is the multiplication depth of a high degree polynomials. The ReLU activation is harder to approximate than GELU. in Sect \ref{sec:backbones} we will provide analysis about choosing the right backbone according to the limited resources in HE.

% The second aspect is regarding skip-connections in NNs. Skip-connections increasing the latency during inference due to \textbf{"chain index mismatch"} \cite{}. In section \ref{sec:skip} we formalize this phenomena and propose a solution to reduce the number of skip connections, that reduce the inference time by a factor of \textcolor{red}{X}.

\paragraph{Our Contributions.}
\begin{enumerate} 
    \item Our main contribution is a novel training method that is grounded by our insight from Sect. \ref{sec:ranges}, which handles the range to the non-polynomial layers. This method enables us to achieve low-degree polynomial approximation, while maintaining the accuracy of the original model.
    
    \item We provide several insights about the design choices of HE-friendly \glspl{CNN}, which can lead to better latency efficiency with a lower approximation error. Specifically, we refer to techniques such as handling neural activations (Sect. \ref{sec:ranges}), \textit{\glspl{SC}} (Sect. \ref{sec:sc}), and the CNN backbone in the context of HE (Sect. \ref{sec:backbones}) .
    
    \item Using the above techniques, we demonstrate, for the first time, the feasibility of training HE-friendly (polynomial) CNNs such as ConvNeXt and ResNet over large scale datasets. These models achieve comparable accuracy to state-of-the-art (SOTA) approaches when trained on realistic datasets like ImageNet  (see Tab. \ref{tab:main}). Our code is available online \footnote{For reproducing our main results, please refer to our anonymous repository: \url{shorturl.at/lvNXZ}. The entire Git repository will be shared upon acceptance.}.
    
    \item We extend the capabilities of employing secure transfer learning over HE. This allows for the first time several key techniques such as encryption of the entire pre-trained model, fine-tuning the entire model rather than optimizing the last layer, and exploiting ZSL as an alternative to training on encrypted data (see Sect. \ref{sec:app}). This demonstration represents a significant milestone in making HE applicable.  

\end{enumerate}

%\paragraph{Our Results.}
\paragraph{Empirical Contributions.}
We implemented and tested our methods using the HElayers framework \cite{helayers}; please see the results in Section \ref{sec:exp}. Consequently, we report the first non-interactive \gls{PPML} solution that can run secure prediction of the above \textbf{large and accurate \glspl{CNN} over large-scale datasets} in minutes, which proves the practicality of secure prediction HE-based solutions. In addition, we take polynomial networks to the next level, by demonstrating the practicality of our approach on the first secure zero-shot and multi-modal foundation model over encrypted data using CLIP (Section \ref{sec:app}).

%\paragraph{Paper Organization.} The rest of the paper is organized as follows: Section \ref{sec:preliminaries} provides background about \gls{HE} and polynomial approximations. Section \ref{sec:method} presents our method for training HE-friendly \glspl{CNN}. We explain how to leverage the proposed method to generate a \gls{ZSL} protocol that can run under \gls{HE} in Section \ref{sec:app} and report our experiments setup and results in Section \ref{sec:exp}. Finally, 
%we conclude the paper in Section~\ref{sec:conc}. A work survey related to this subject is shown in App. \ref{sec:rel}. 

%To address these issues, non-polynomial operations, such as tanh, max, and division operations, have been added to stabilize the training process, making the NN not polynomial.

\section{Background}\label{sec:background}

%Training deep \glspl{CNN} from scratch is a difficult problem~\cite{goyal2020improved}, and often does not achieve the same level of performance as the original network. Although some previous studies have achieved promising results on small \glspl{CNN} such as AlexNet \cite{baruch2021fighting} or small datasets such as CIFAR-10/100 \cite{pmlr-v162-lee22e}, we found that their methods could not scale up for larger networks like ResNet and ConvNeXt, even with distillation techniques. 

To clarify the difficulty of producing polynomial networks several theoretic intuitions and proofs were proposed. For example, \cite{zhou2019polynomial} proved that under some conditions polynomial \glspl{FFN} are unstable, and concluded that the more complicated a polynomial activation is the more likely that it will face instability. Another paper, \cite{goyal2020improved}, suggests that the problem with polynomial activations is that the gradients and outputs are unbounded and can be arbitrarily large, in contrast to other activations such as ReLU, GELU, Sigmoid, or TanH. 
The paper also points out that in deeper networks $f_{(d,l)}$ with $l$ layers and $d$-degree polynomial activations, the gradients explode exponentially in the degree of the entire network, since for input $ x > 1, \lim_{x \rightarrow \infty} f_{(n.l)}(x)/x=\infty$. Additionally, \cite{chrysos2020p}, \cite{goyal2020improved} and \cite{gottemukkula2020polynomial} attempted to implement deep polynomial networks but faced optimization instability. They resolved the issue by incorporating non-polynomial components like tanh or max, resulting in a non-polynomial model.%Moreover, polynomial NNs are also limited in their expressiveness \cite{???}.

\paragraph{Polynomial Approximations.} Instead of training a polynomial network from scratch, a commonly used method is approximating non-polynomial functions of pre-trained networks using polynomials. For example, the \ReLU activation function is approximated by a polynomial in the studies of  \cite{relu1, relu2, SecureML, cryptoDL} or is replaced by a trainable polynomial in \cite{baruch2021fighting}. One commonly used way to approximate a function is by using the well-known Remez algorithm \cite{Remez} and its follow-up algorithms \cite{RI1, RI2}, which were proved to be optimal tools for finding the polynomial approximation of a function $f(x)$ given the range of $x$ and polynomial degree. Nevertheless, the range of the different CNN layers' input $x$ may not be known in advance, which may lead to a non-negligible error due to the approximation's poor performance outside of the conditional range. See more details in Appendix. \ref{sec:polapporx}. 
One interesting work is \cite{lee2021precise} that approximated \ReLU using a composition of 3 polynomials of degrees \{15, 27, 29\}. While the reported accuracy was only 0.11\% lower than \gls{SOTA}, the authors of \cite{lee2021precise} did not test their approach in a low-precision environment such as \gls{HE}.%
%\textcolor{red}{ TODO:explain that their method generates polynomial of extreme degrees that cannot be ran on imagenet}

\begin{figure}[t!]
    \centering
    \begin{minipage}[b]{0.42\linewidth}
        \centering
        \includegraphics[width=\linewidth]{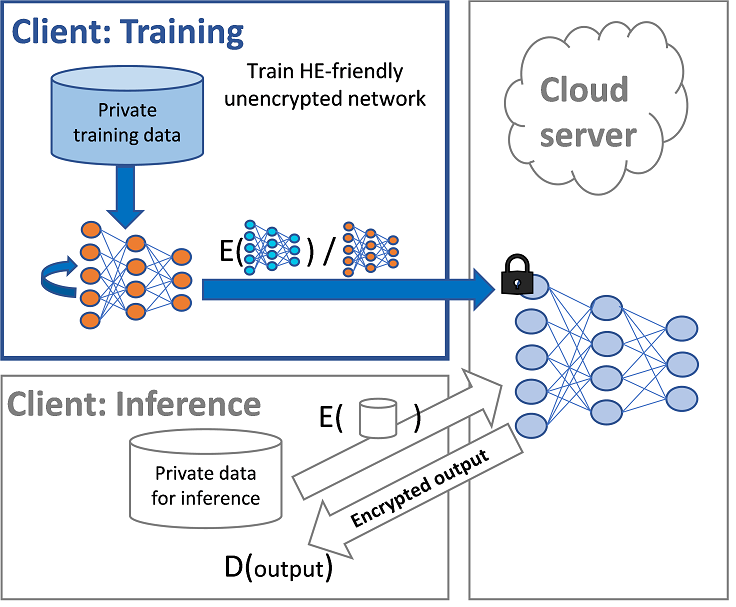}
        \caption{\textbf{(Motivation)} An E2E \gls{PPML} solution for running \glspl{CNN} over \gls{HE}. The flow involves a client and a cloud server. The client \textbf{trains a polynomial (\gls{HE}-friendly) \gls{CNN} model}, either encrypts it or not, and uploads the model to the cloud. Then, the client requests from the cloud to run this model on its behalf. For that, the client encrypts its private samples and uploads them to the cloud, which \textbf{processes the encrypted data using the (possibly encrypted) model}, and returns the results to the client for decryption.}
        \label{fig:figure1}
    \end{minipage}
    \hfill
    \begin{minipage}[b]{0.56\linewidth}
        \centering
        \begin{minipage}[b]{0.95\linewidth}
            \centering
            \includegraphics[width=0.98\linewidth]{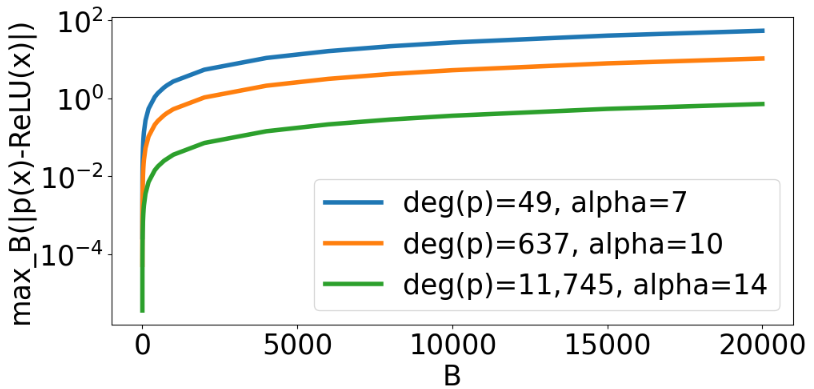}
            \caption{\textbf{(Problem)} Maximal error when using the $p_{\alpha=7,10,14}$ polynomials of \cite{lee2021precise} to approximate \ReLU over different ranges ($B$). Small error is achieved through a small range or a large polynomial degree.}
            \label{fig:MiniMaxErrorperRange}
        \end{minipage}
        \begin{minipage}[b]{0.95\linewidth}
            \centering
            \includegraphics[height=4cm, width=0.98\linewidth]{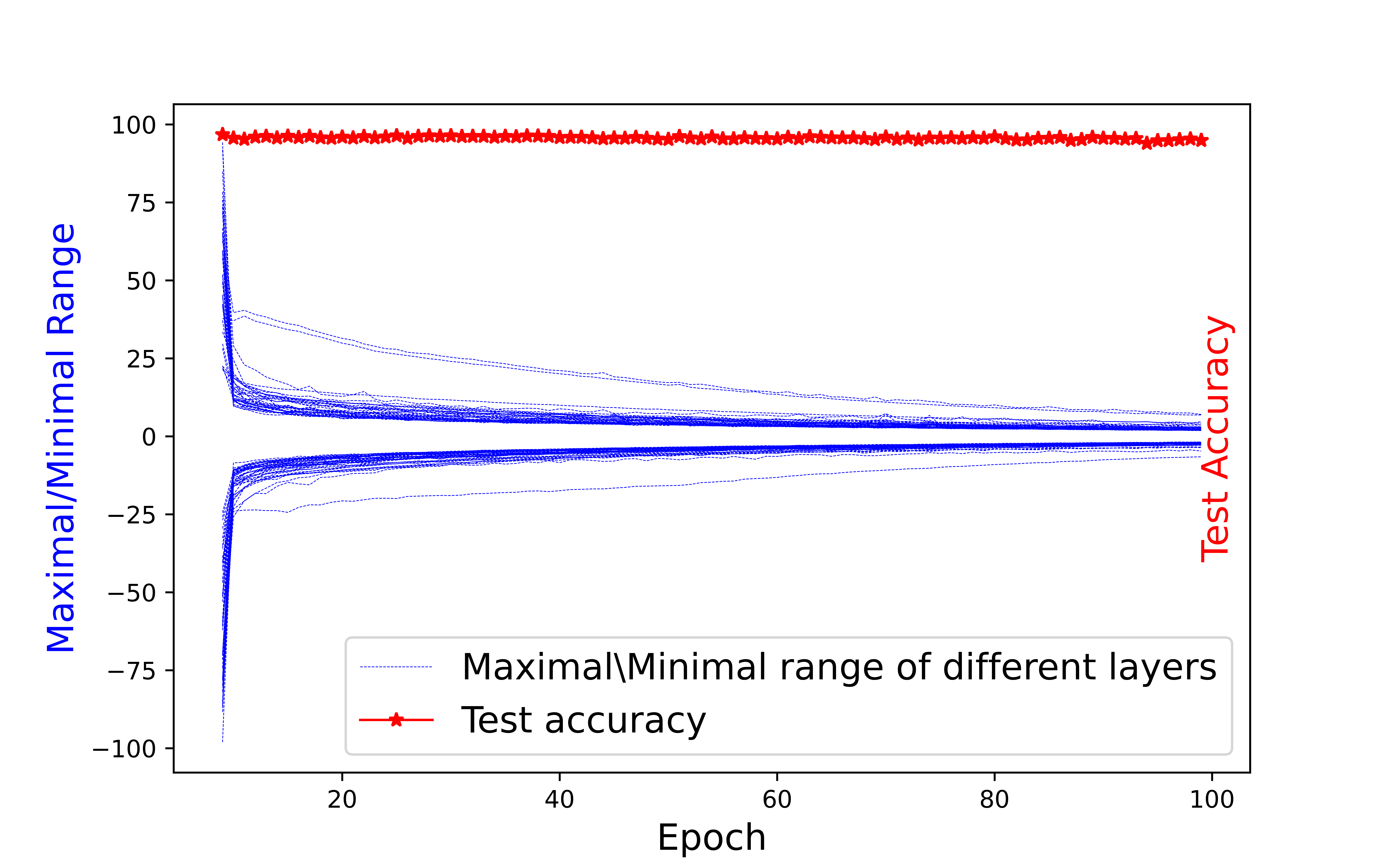}
            
            \caption{\textbf{(Solution)} range-aware training. Here, accuracy and ranges of ConvNeXt trained on CIFAR-10.}
            \label{fig:range_loss_training}
        \end{minipage}
    \end{minipage}
\end{figure}

\paragraph{Motivation for training polynomial networks.}
Our principal motivation in this work is to enable E2E-secure inference (in contrast to client-aided based solutions, see \ref{paragraph:client_aided}) that uses \gls{HE}. This will allow data owners to use third-party cloud environments while complying with regulations such as GDPR \cite{GDPR} and HIPAA \cite{HIPAA}. An example of a problem-setting is provided in Fig. \ref{fig:figure1}. For brevity, we only claim that HE-friendly CNNs should be polynomial and refer the interested reader to App. \ref{app:he} for more details about \gls{HE}. 

% Our principal motivation in this work is to enable secure inference that uses \gls{HE}, this will allow data owners to use third-party cloud environments while complying with regulations such as GDPR \cite{GDPR} and HIPAA \cite{HIPAA}. An example of a problem-setting is provided in Fig. \ref{fig:e2eflow}. For brevity, we only claim that HE-friendly CNNs should be polynomial and refer the interested reader to App. \ref{app:he} for more details about \gls{HE}. 

\paragraph{Related art.}
HE-based secure prediction solutions should be both efficient and accurate. In App. \ref{sec:rel} we provide a detailed comparison of \gls{SOTA} HE-based \gls{PPML} solutions. Here, we only summarize that the most efficient solutions today \cite{pmlr-v162-lee22e, hyphen} both reported accuracy only for CIFAR-10/100, where \cite{hyphen} mentions that they have not yet succeeded in training ResNet-18 over ImageNet. In contrast, as mentioned above, the most accurate attempt to run an HE-friendly inference is of \cite{lee2021precise} who did not implement their solution with HE. Followup works e.g., \cite{sisyphus} claimed that due to the large polynomial degree (e.g., more than 10K), the solution latency when evaluated under HE is large, and \cite{resnet20} only showed practicality under HE for ResNet-20 and CIFAR-10. Furthermore, our experiments (see supplementary material) were not able to reproduce the results of \cite{lee2021precise} for ResNet-50/150 over ImageNet, even when using 96-bit floating-point precision on plaintext. % It seems that achieving both an accurate and efficient solution is challenging. %
In conclusion, our work provides the first accurate and performant implementation of large polynomial CNNs on large datasets.

%{\color{red} For example, the authors of \cite{lee2021precise} who demonstrated an implementation of ResNet-20 on CIFAR-10 used a composition of three polynomials $P=P_1(P_2(P_3(x)))$ of degrees $\{15,27,29\}$, respectively, to approximate the ReLU function. The final degree of $P$ is therefore $11{,}745$ and the multiplication depth is $\ceil{\log_2(6{,}075))}=14$}
% One potential solution is the utilization of \gls{HE}, which enables computation on encrypted data. See App. \ref{app:he} to learn more about \gls{HE} and App. \ref{sec:rel} for a review and a comparison of \gls{SOTA} HE-based \gls{PPML} solutions. 

% Studying the methods for generating these HE-friendly \glspl{CNN} is the main focus of this paper. This can be done by replacing the non-polynomial components with a polynomial similar operation that is not a direct approximation of the original layer. A good example of this is replacing a max-pooling operation with mean-pooling, which in many use cases does not affect the \gls{CNN} performance \cite{CryptoNets2016}, while in other use cases it might introduce too much noise \cite{baruch2021fighting}. The other approach is using polynomial approximations  \cite{relu1, relu2, SecureML, cryptoDL}. 

\section{Training Method}\label{sec:method}
Our method for generating practical HE-friendly CNNs is based on three essential factors: a) Minimizing input ranges to activation layers and generating suitable polynomial approximations based on these ranges  (Sect. \ref{sec:ranges}); b) Effectively handling of \glspl{SC} under HE (Sect. \ref{sec:sc}); and c) Choosing the appropriate backbone architecture (Sect. \ref{sec:backbones}). 

% \subsection{Handling Input Ranges for Polynomial Approximations}\label{sec:ranges}
\subsection{Input Range Tuning for Accurate Polynomial Approximation of Activation Functions\label{sec:ranges}}

Approximating activation functions with polynomials accurately and efficiently, and applying them in deep \glspl{CNN} is a hard task. We identified that the main issue in approximating non-polynomial networks is that the input range for these polynomial approximations is not known in advance, often spanning over scale of hundreds \cite{lee2021precise}, and thus the deviation of the original activation from the approximated activation increases (see Fig. \ref{fig:MiniMaxErrorperRange}, Observation \ref{obs:range}). In practice, when dealing with large networks with multiple approximated layers, the error from the initial layers is accumulated and eventually causes explosions and instability. %the network to ``explode'', i.e., the output value of the polynomial activation functions becomes too high for the network to converge.

Traditionally, to reduce the approximation error and thus the accumulated error, the network designer is forced to use high-degree polynomials \cite{lee2021precise}. %For example, the authors of \cite{lee2021precise} who demonstrated an implementation of ResNet-20 on CIFAR-10 used a composition of three polynomials $P=P_1(P_2(P_3(x)))$ of degrees $\{15,27,29\}$, respectively, to approximate the ReLU function. The final degree of $P$ is therefore $11{,}745$ and the multiplication depth is $\ceil{\log_2(6{,}075))}=14$.
However, as detailed below, we take a different approach, in which we reduce the polynomial degree by reducing the input range for every polynomial. This reduces the accumulated error and hence the chances of a network explosion to occur.%

Alg. \ref{alg:method} provides a high-level overview of our method. 
%It receives a pre-trained \gls{CNN} model \M, %a training set (\trainset), a small disjoint set (\rangeset), and a degree $d$ as input, where \trainset and \rangeset together form the training set of \M.
Let \M be a pre-trained non-polynomial model, \NPL be the ordered list of length $\NPLSize$, that contains the non-polynomial layers of \M. %$\NPLSize=|\NPL|$ be its size.%
Let $c_i = |\NPL[i]|$ be the number of neurons at layer $i$ in \NPL, $\mathbf{x}^i$ be the vector input for that layer and $d$ be the polynomial degree.

\paragraph{The first phase} of the algorithm involves adding a novel regularization term, \textit{range loss} (\regterm), to \M's original objective function. This loss term aims to reduce the range of inputs to the \NPL layers around the value of 0, which can be depicted as 
$rl = \lVert ( \lVert \mathbf{x}^i\rVert_p)_{0 \le i < \NPLSize}\rVert_q$, where we often set $p=\infty$ and $q \in \{1, 2, \infty\}$.
The new loss function for input $(X, y)$ is defined as: $loss(\M) = CE(\M(X), y) + w \cdot \regterm$, where CE is the Cross Entropy loss on the model. %, though any other loss function can be used for a specific model.

When using the $L_1$ norm for \regterm and when the size of \NPL increases, the range loss term may become more significant than the original CE loss, which is why we introduce a weight $w$ to balance the two terms. In Step 2, the algorithm fine-tunes \M using the new loss function. This phase ends when the loss is minimized, at which point the activation functions should expect inputs in the minimal range so that the model preserves its performance. Fig. \ref{fig:range_loss_training} demonstrate the effectiveness of this
%fine-tuning 
procedure.  

\paragraph{In the second phase of Alg}. \ref{alg:method} (Steps 3-4), the framework uses \textbf{empirical analysis} to estimate the input ranges per layer $[x^i_{min} < x^i_{max}]_{0 \le i < \NPLSize}$, with confidence level $\alpha$. This is done by approximating the ranges of the input to each activation function by sampling a subset of the training data that has not been used for training or validation. This stage takes into account the error generated by the approximation of the previous layer, i.e., it pre-bounds the error with $e_i$ and expects that in the last step, the approximation would be bounded by $|p_i(x) -
f_i(x)|< e_i$.

\begin{algorithm}[b!]
\caption{Training HE-friendly \glspl{CNN}}
\label{alg:method}
\begin{algorithmic}
    \STATE {\bfseries Input:} A pre-trained \gls{CNN} model (\M), a training set (\trainset), a small disjoint set (\rangeset), and a positive integer degree ($d$).
    \STATE {\bfseries Output:} A trained HE-friendly model $(\M_{HE-f})$.
\end{algorithmic}
\begin{algorithmic}[1]
    \STATE Add a regularization range loss term $rl$ to loss(\M).
    \STATE Fine-tune \M over \trainset until the input ranges to the \NPL layers are small enough and the network performance is satisfying. The resulting model is \M'.
    \STATE Evaluate \M' over \rangeset and compute the pairs $(\min{\mathbf{x}^i}, \max{\mathbf{x}^i})_{0 \le i < \NPLSize}$ per sample.
    \STATE Using the above pairs, estimate the range $([x^i_{min}, x^i_{max}])_{0 \le i < \NPLSize}$ for values of $\mathbf{x}^i$ with confidence level $\alpha$. 
    \STATE Replace the functions $f_i(x)$ of the \NPL layers with polynomial approximations $P_i(x)$ of degree $d$ over the estimated ranges $[x^i_{min}, x^i_{max}]$. The new model is $\M_{HE-f}$.
    \STATE Fine-tune $\M_{HE-f}$ over \trainset until convergence. 
    \STATE return $\M_{HE-f}$.
\end{algorithmic}
\end{algorithm}

\paragraph{In the last phase} (Steps 5-6), we replace the original activation functions with polynomial approximations, using e.g., Remez or the faster but less accurate least-square polynomial fit function. Each activation layer is replaced by a separate polynomial that has been designed for the estimated range. The output of the algorithm is an HE-friendly model $\M_{HE-f}$. However, since the polynomial activation layers provide only an approximation of the original activations, the accuracy of the model is normally decreased. Therefore, we added Step 7 to fine-tune the model for a few more epochs with the added \regterm term until the desired performance is achieved.

\begin{figure*}[ht!]
    \centering
    \begin{subfigure}[b]{0.24\textwidth}
        \centering
        \includegraphics[width=0.98\textwidth]{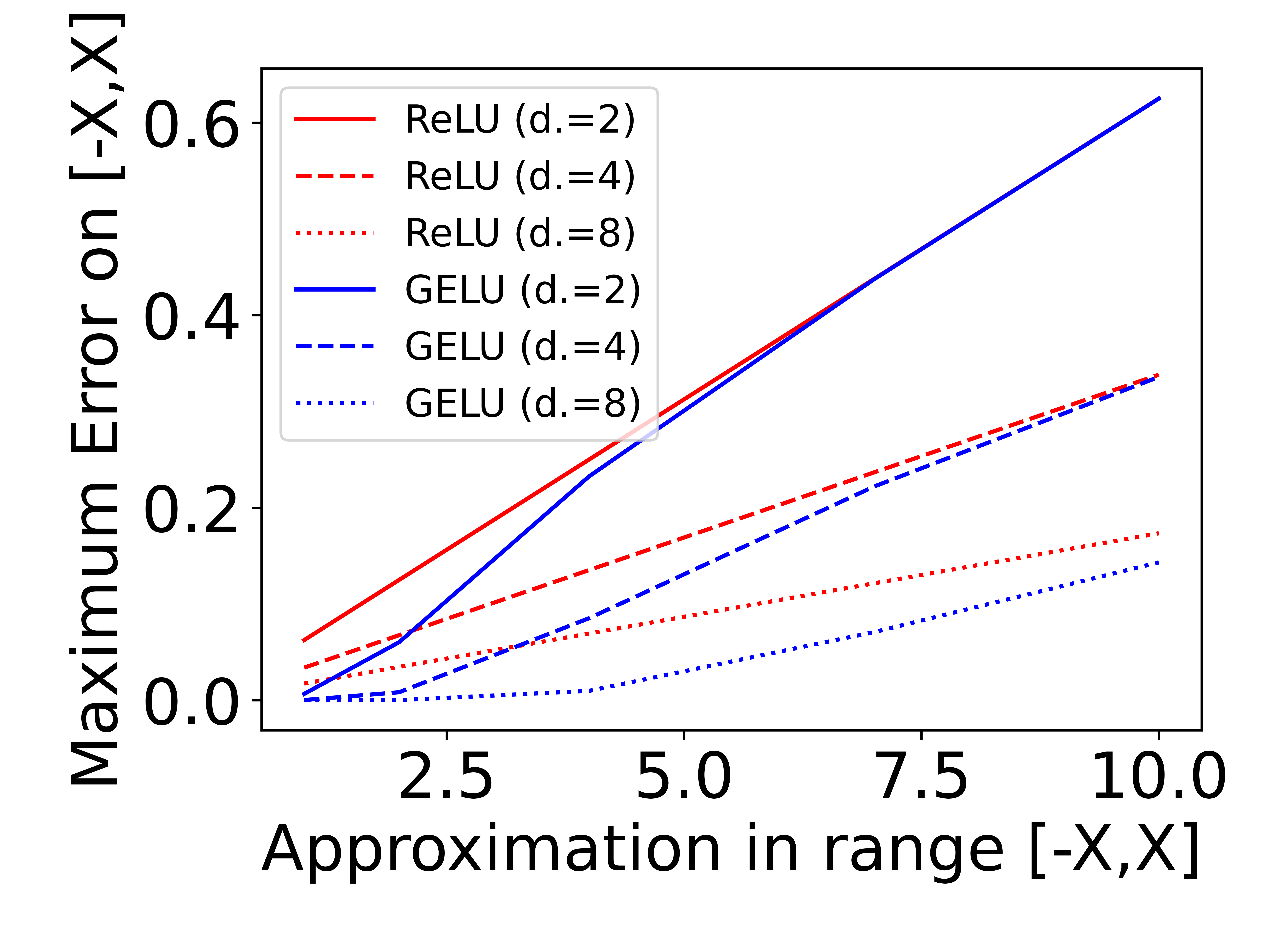}
        \caption{Poly. approx. error}
    \end{subfigure}
    \begin{subfigure}[b]{0.24\textwidth}
        \centering
        \includegraphics[width=0.98\textwidth]{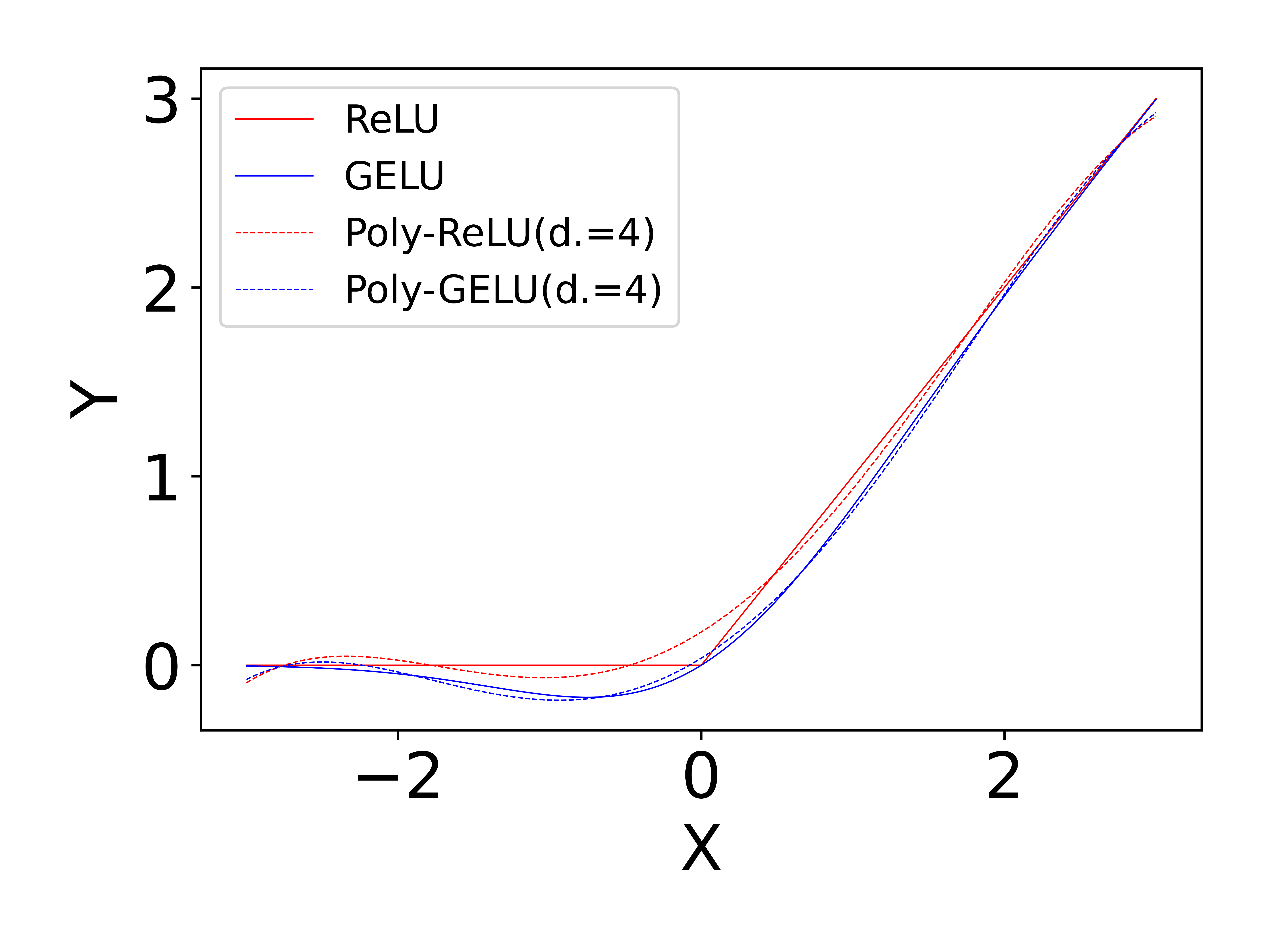}
        \caption{Approx. polynomials}
    \end{subfigure}
    \begin{subfigure}[b]{0.24\textwidth}
        \centering
        \includegraphics[width=0.98\textwidth]{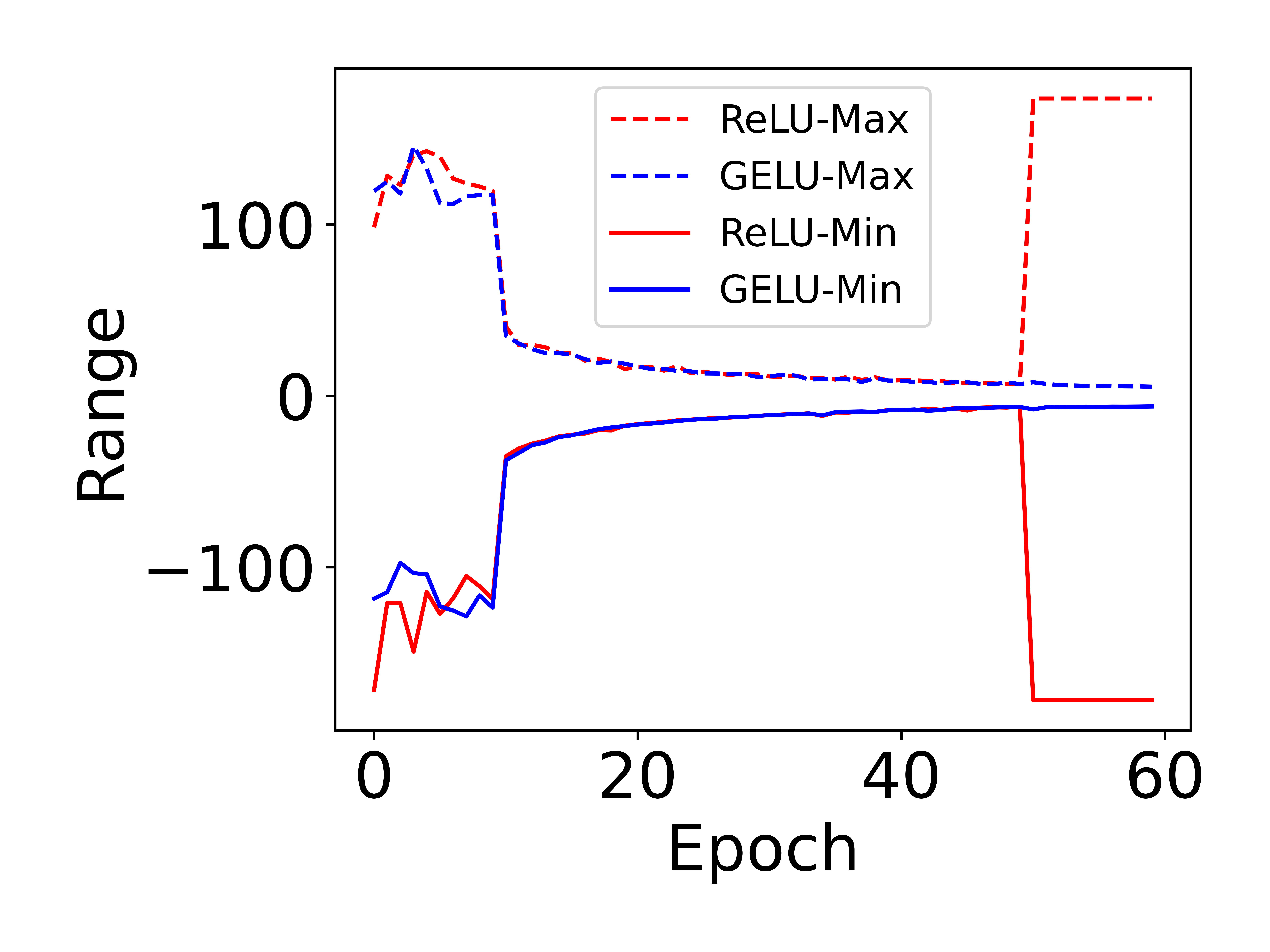}
        \caption{Min/max ranges/epoch}
    \end{subfigure}
    \begin{subfigure}[b]{0.24\textwidth}
        \centering
        \includegraphics[width=0.98\textwidth]{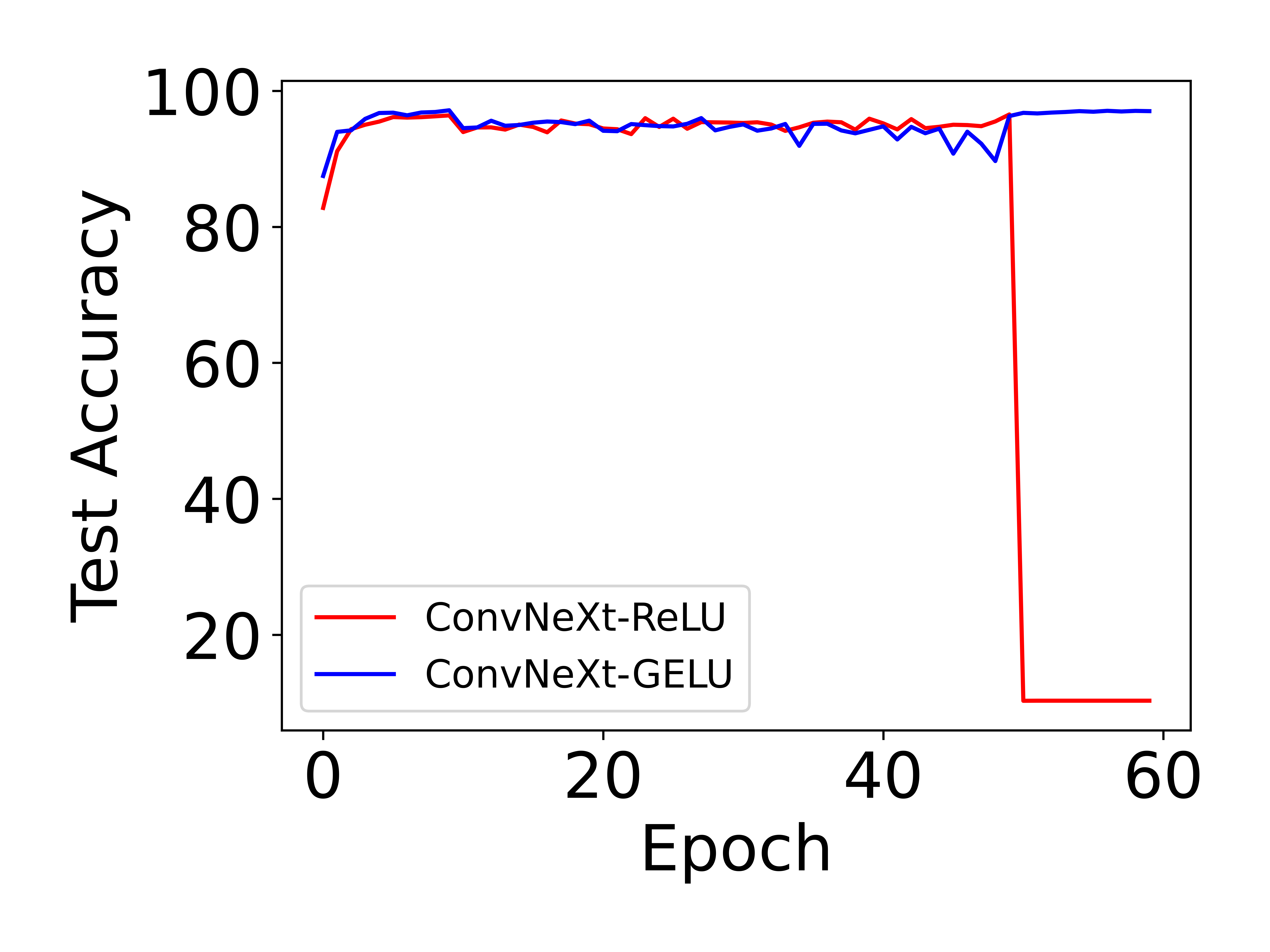}
        \caption{Test accuracy epoch}
    \end{subfigure}
    \caption{\ReLU (red) versus \GeLU (blue). Panel (a): Maximum error $|p(x)-f(x)|$ for a different degree of polynomial approximation $p(x)$ of \ReLU/\GeLU in different ranges (x-axis). Panel (b): A 4- degree polynomial approximation of \ReLU/\GeLU. In both (a) and (b), \GeLU is better approximated. Panel (c) Error range [min, max] (y-axis) after $l$ training epochs (x-axis). Panel (d): Model accuracy (y-axis) after training ConvNeXt with \ReLU/\GeLU for (x-axis) epochs: In the initial 10 epochs, max-pooling and LayerNorm are substituted with HE-friendly components (mean-pooling and BatchNorm). Our range-aware training technique is then applied from epochs 10 to 50. Finally, at epoch 50, the activations are replaced by polynomials. While the ranges exhibit similarity, only \GeLU can be precisely approximated.%. In all cases, \GeLU performs better than \ReLU. 
    }
    \label{fig:compareActivations}
    \vspace{-10pt}
\end{figure*}

\subsection{Efficiently Handling Skip-Connections in HE} \label{sec:sc}
CKKS and other HE schemes have a limit on the number of multiplications that can be performed on a ciphertext, known as the "multiplication chain index'' (a.k.a. modulus chain index) or \ChainIdx. This limit is set by the client to achieve the desired level of security and performance \cite{standard}. Every ciphertext starts with a \ChainIdx of $0$. After each multiplication of two ciphertexts with \ChainIdx of $x$ and $y$, the result has a \ChainIdx of $\max(x,y) + 1$. Once a ciphertext's \ChainIdx reaches the limit, it can no longer be multiplied, unless a costly $\Bootstrap$ operation is performed to reduce its \ChainIdx, or even reset it back to $0$. One design goal when generating an HE-based solution is to minimize the number of $\Bootstrap$ invocations. Hereafter, we define the term \textit{multiplication depth} to be the longest chain of sequential multiplication operations in an \gls{HE}-evaluated function. As noted above, longer chains, i.e., higher multiplication depths, result in more bootstrapping operations.
Using these definitions, we observe that
\begin{observation}\label{obs:sc}
Given a \acrlong{SC} layer $SC_f(x) = x + f(x)$, where $f$ is a combination of some layers. When running under HE, $\ChainIdx\left(SC_f(x)\right) \in \{\ChainIdx(x), \ChainIdx(f(x))\}$ and when $\ChainIdx(x) \neq \ChainIdx(f(x))$ the \gls{SC} implementation may increase the overall multiplication depth of the network by $|\ChainIdx(x) - \ChainIdx(f(x))|$.
\end{observation}

In practice, the cost of $SC_f(x)$ can be even higher because the input $x$ or $f(x)$ may need to go through some transformation before adding it to $f(x)$. This is the case with the HElayers SDK, which requires input to an $\operatorname{ADD}$ operator to use the same ciphertext parameters. Specifically, it applies transformations $g,h$ on $x$, $f(x)$, respectively, and replaces $S_f(x)$ with the operator $S_{f,g,h}(x)=g(x)+h(f(x))$. In this case, for Observation \ref{obs:sc}, we need to consider $\ChainIdx(g(x))$ instead of $\ChainIdx(x)$ and $\ChainIdx(h(f(x)))$ instead of $\ChainIdx(f(x))$.

Given the latency costs associated with implementing \glspl{SC} under HE, we propose two methods for placement and removal of \glspl{SC}, where our goal is to maintain accuracy while improving latency. Note that this was not required previously for commonly used networks where addition is free. However, with our methods, we can offer new network designs that are more suitable for the HE world.  

\paragraph{Removing Skip-Connections.}
Our first method aims to generate a \gls{CNN} without \glspl{SC} while maintaining near-\gls{SOTA} performance. Removing \glspl{SC} directly will result in bad performance as previously demonstrated in the study of He et al. \cite{resnet} due to the gradient flow across layers. Therefore, in our method, we start by training a \gls{CNN} to achieve \gls{SOTA} performance while using \glspl{SC}. We gradually eliminate them by replacing $S_{f,g,h}(x)$ with a new layer $S'_{f,g,h, a}(x) = a \cdot g(x) + h(f(x))$. We continue training for $N$ more epochs. At every epoch index $ 0 < E \le N$, we set $a=(1-\frac{E}{N})$. After $N$ epochs, $a=0$ and the \glspl{SC} are removed. Finally, we continue training for several more epochs. 
%Empirical analysis of this method is presented in Tab. \ref{tab:skiptiming}.

\paragraph{Skip-Connection Placement.}

When \glspl{SC} are required, we propose Alg. \ref{alg:scplacement} for designing  efficient HE-friendly \glspl{CNN}.
%(see discussion in Section \ref{sec:discussion}) 
The algorithm uses as input a network architecture \A and an HE network analyzer (\analyzer), such as the optimizer of HElayers. We start by removing all \gls{SC} layers and feeding the new network \A' to the analyzer. For every pair of layers in the network $i < j$, the analyzer computes the latency costs of adjusting the output $x$ and $f(x)$ of layers $i$, $j$, using transformations $g,h$, respectively, and the need for \Bootstrap operations. It returns the results in the matrix \costsMatrix of size $L \times L$, where $L$ is the number of layers in \A'. For $i \ge j$, $\costsMatrix[i][j] = \infty$. Using \costsMatrix, we can now place \glspl{SC} to minimize the latency overhead while maintaining accuracy. One possible heuristic is to start from layer $i=0$ and find $j = \argmin_j{\left(\costsMatrix[i][j]\right)}$, place an \gls{SC} between layers $i$ and $j$, and repeat with $i=j+1$. Note that when adding \gls{SC} in a way that increases $\ChainIdx(h(f(x)))$ the costs matrix should be re-computed.
Alg. \ref{alg:scplacement} already considers the bootstrapping costs discussed earlier, and is most likely to choose \glspl{SC} for layers with $\ChainIdx(g(x)) \le \ChainIdx(h(f(x)))$.

\begin{table}[t!]
    \centering
    \begin{minipage}[t]{0.48\textwidth}
        \begin{figure}[H]
            \centering
            \includegraphics[width=1\linewidth]{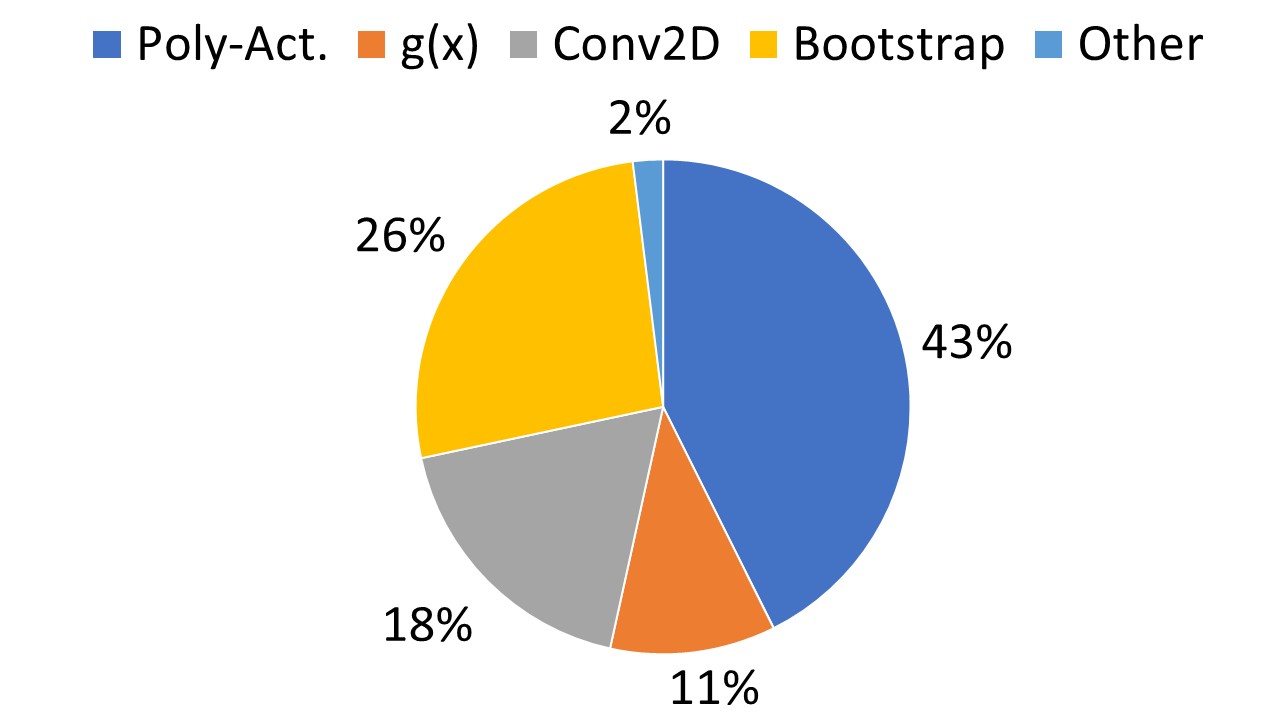}
            \caption{Latency breakdown for running ResNet-50 over ImageNet using HElayers 1.5.2. Here, $g(x)$ is the \gls{SC} adaption of HElayers. \textit{Other} refers to BN, FC, and AVG-Pool layers.}
            \label{fig:profiling}
        \end{figure}
    \end{minipage}\hfill
    \begin{minipage}[t]{0.48\textwidth}
        \centering
        \begin{algorithm}[H]
        
        \caption{\gls{SC} HE-friendly design}
        \label{alg:scplacement}
        \begin{algorithmic}
            \item[]
            \STATE {\bfseries Input:} A network architecture (\A) and an HE analyzer \analyzer.
            \STATE {\bfseries Output:} An HE-friendly NN architecture \A'.
        \end{algorithmic}
        \begin{algorithmic}[1]
            \item[]
            \STATE Set \A' to be \A without \glspl{SC}.
            \STATE $costsMatrix_{L,L} = \analyzer(\A')$
            \STATE Place \glspl{SC} in \A' between layers $i,j$ considering the latency costs $costsMatrix[i,j]$ until reaching the desired accuracy.
            \STATE Return \A'
            \item[]
          %  \item[]
        \end{algorithmic}
        \end{algorithm}
    \end{minipage}\hfill
    \vspace{-20pt}
\end{table}

\vspace{-4pt}
\subsection{Choosing the Right Backbone}\label{sec:backbones}
Many variations of ResNet backbones have been proposed over the years \cite{convnext1,resnext,wideresnet,densenet,resnet_in_resnet}. While most prior HE-related works use vanilla ResNet \cite{pmlr-v162-lee22e, lee2021precise, autoFHE} (see App. \ref{sec:rel}), its superiority has not been proven or explored in the context of HE. Surprisingly, even seemingly insignificant design choices such as the backbone choice can have a large performance impact when working with encrypted data. Therefore, we decided to study and compare three ResNet backbones: the original ResNet~\cite{resnet}, ConvNeXt~\cite{convnext1}, and DenseNet~\cite{densenet}. 
% We evaluate them based on two criteria: 1) the overall accuracy of the HE-friendly CNN, 2) the latency when executed under HE.

%\subsubsection{ConvNeXt.}
\paragraph{ConvNeXt.} ConvNeXt is a modern variant of ResNet, which achieves the highest performance \cite{convnext2}. It involves two relatively minor design decisions that make it attractive when working over encrypted data: 1) it has a reduced number of activations; 2) it uses \GeLU.

%\textbf{Reduced number of activations.} 
Every ConvNeXt block includes only one activation function, as opposed to three in ResNet blocks. Although this only provides a relatively small improvement of ($0.7\%$), it has crucial implications in HE since smaller numbers of non-polynomial components reduce the overall multiplication depth by a factor of around $3$, making it easier to achieve an HE-friendly network. 

%\textbf{Using GELU.} 
ConvNeXt uses the \GeLU~\cite{gelu} activation function instead of \ReLU. Fig. \ref{fig:compareActivations} Panels (a) and (b) show that \ReLU is much more difficult to approximate in comparison with \GeLU, specifically in the areas near 0, where \ReLU is not smooth. Panels (c) and (d) show that using \GeLU is more robust, allowing us to reduce the polynomials' degree. The result is that the overall replacement process is easier, and the aggregated multiplication depth is smaller.

% Since \gls{SC} is a bottleneck in HE, architectures such as DenseNet significantly increases bootstrapping operations, making it unsuitable for HE-based PPML solutions.
%\subsubsection{DenseNet.}
\paragraph{DenseNet.}
The DenseNet network uses a unique type of \glspl{SC}. Each layer receives the outputs of all preceding layers as inputs, and its output is used as input for all subsequent layers. While this architecture has advantages in terms of optimization, it increases the number of bootstrapping operations, since it forces bootstrapping after almost every DenseNet block. Thus, it is not recommended for \gls{HE}-based \gls{PPML} solutions.

\section{Experiments and Results}\label{sec:exp}

To test our methods, we performed a series of experiments, which we report next. For brevity, we refer to ResNet-XX as RNXX. All training experiments used the PyTorch framework.

\paragraph{HE-Friendly large-scale CNNs.}
We start by evaluating the accuracy of our proposed training method for HE-friendly models on three datasets: CIFAR-10 \cite{CIFARDatasets} and the large-scale datasets: ImageNet \cite{deng2009imagenet} and Places-365 \cite{zhou2017places}. All datasets were evaluated at a resolution of $224\times 224\times 3$. The accuracy results, presented in Tab. \ref{tab:main}, were analyzed in three stages: (1) the original, non-HE-friendly model, (2) the original model after replacing max-pooling with mean-pooling but with non-polynomial activations, and (3) the proposed HE-friendly model with polynomial activations. Additionally, we compared our results to those reported by \cite{pmlr-v162-lee22e}, who evaluated a ResNet-56 model on CIFAR-10 using images of size $32 \times 32 \times 3$. While they also report on other networks, we compared our results to their ResNet-56 as it is the closest to ResNet-50. The smaller image size used by Lee et al. \cite{pmlr-v162-lee22e} may have contributed to the lower accuracy observed in their experiments.

During the range minimization phase (Alg. \ref{alg:method}, Step 2), the ranges to the activation layers are large, which leads to the \regterm loss being significantly higher than the original model loss. To this end, we set $w$ to be in the ranges 0.0001-0.001 and 0.01-0.1, before (Step 2) and after (Step 5), resp., replacing the \ReLU activation with a polynomial, the exact value depends on the dataset. The polynomial degree used in these experiments is set to 18, which provides a good approximation for small ranges of around $[-10,10]$, as in our case. Tab. \ref{tab:main} shows that the HE-friendly models trained by our method preserved the original accuracy when applied on CIFAR-10 and Places-365, and reached 94\% accuracy when trained on ImageNet for ResNet-101 and 96\% for ConvNext-T.

\paragraph{HE-Friendly Skip-Connections.}
Tab. \ref{tab:skiptiming} demonstrates the effect of using  our method of carefully removing \glspl{SC} on secure classification latency. For that, we use ResNet-50 and set the polynomial activation degree to be 2, 8, 16 or 18. We see that using HE-friendly skip-less models can save up to $75\%$ of the bootstrapping as well as provide significant speedup. Recall that bootstrapping is a critical bottleneck in inference under HE, see App. \ref{sec:profiling}.

When it comes to accuracy, the situation is more complicated. We used our training method with only $18$-degree polynomial activation functions. We found that accuracy degradation is dependent on whether the network starts with pre-trained weights. Without pre-training, we observe that HE-friendly ResNet-50 and skip-less HE-friendly ResNet-50 achieve $93.72\%$ and $93.21\%$, resp., where the total degradation is relatively small ($0.53\%$). In contrast, when using a pre-trained model, the impact on accuracy is more significant, see Tab. \ref{tab:results-backbone}. This phenomenon is somewhat predictable, since removing the \glspl{SC} changes the network's dynamics, thus diminishing the impact of pre-training data.

\paragraph{HE-Friendly Backbones.}
We tested different settings for CNNs, including various sizes of ResNets (18, 50 ,101), ResNet-50 without \glspl{SC}, ResNet-50 with adaptive removal of \glspl{SC}, and two variations of ConvNeXt-Tiny, which is equivalent to ResNet-50: one with a polynomial degree of 4, and the other with a polynomial degree of 8. All of the experiments were applied on CIFAR-10 at a resolution of  $224\times 224\times 3$. Results are detailed in Tab. \ref{tab:results-backbone}. 
We find that the ConvNeXt-Tiny model with $4$-degree polynomial activations has a significantly lower multiplication depth compared to the ResNet-50 model with 18-degree polynomial activations. Despite similar numbers of blocks, FLOPs, and accuracy, the number of \Bootstrap operations in ConvNeXt-Tiny is reduced from $7{,}712$ to $1{,}360$. This improvement is significant, as the bootstrap operation is a major bottleneck in the inference time of deep CNNs, as shown in the profiling provided in App. \ref{sec:profiling}. Results are provided in Tab. \ref{tab:results-backbone}. When implementing an HE-friendly ConvNeXt, we replaced the LayerNorm layers with BatchNorm layers, which resulted in some degradation of accuracy. We also assume that the replacement also negatively impacts the model’s transfer-learning capabilities. %
% Temp for submit draft
For a comprehensive overview of the experimental setup we use for evaluation over HE via HElayers SDK, please see Appendix \ref{sec:hesetup}
% For a comprehensive overview of the experimental setup we use for evaluation over HE via HElayers SDK, please see Appendix \ref{sec:hesetup}
% 

% \paragraph{Performance of Predictions over Encrypted Data.}
% For the following experiments, the models running time is reported for both the GPU and the CPU hardware, which we describe in App. \ref{sec:hesetup}. The GPU runs were based on an average of $500$ samples per dataset, while the CPU runs were based on a smaller sample size of $10$ due to the longer processing time. In addition, we evaluate the accuracy of the model when applied to encrypted and unencrypted data, and reach an MSE in the range $[1e^{-12}, 1e^{-10}]$. We evaluated our models under HE using the HElayers SDK. Tab. \ref{tab:models} summarizes the latency and memory results of ResNet-18, 50, 101 and 152 with 128-bit security.

%Tab. \ref{tab:models} summarizes the latency and memory results.

 \begin{table}[t!]
    \centering
    \begin{minipage}[t]{0.48\textwidth}
         \begin{table}[H]
            \footnotesize
            \centering
            \caption{Accuracy of HE-friendly backbones over \textbf{CIFAR-10} $(224\times 224\times 3)$.}
            \label{tab:results-backbone}
            \begin{tabular}{|l|c|c|c|}
                \hline
                \textbf{Arch.} & \textbf{Original} & \textbf{HE-friendly} \\
                \hline
                RN18            &  99.36       &  99.3   \\
                RN50            &  99.58       &  99.5    \\
                RN101           &  99.8        &  99.58     
                \\
                RN152          & 99.9  & 99.6 \\ \hline
                RN50 no skip    &            93.5                  & 93.58 \\ \hline
                %RN50 adaptive   &            95.27                    & \todo{94.12  \\ \hline
                ConvNext-Tiny$_{d=4}$         &  97.84                  &  97.03     \\
                ConvNext-Tiny$_{d=8}$         &  97.84                  &  97.34     \\
                 RN50$_{32\times32}$ &    97.8                    &  97.0 \\
                 RN56-\cite{pmlr-v162-lee22e}$_{32\times32}$ &    97.8          &   93.27          \\ 
                \hline
        \end{tabular}
        \end{table}
    \end{minipage}\hfill
    \begin{minipage}[t]{0.48\textwidth}
        \begin{table}[H]
            \footnotesize
            \centering
            \caption{Accuracy results of training ResNet (RN) and ConvNeXt-Tiny (CNXT) over different datasets. The results include the original and our HE-friendly model with poly. activations.} 
            \label{tab:main}
            \begin{tabular}{|l|l|c|c|}
                \hline
                 \textbf{Arch.} & \textbf{Dataset} & \textbf{Original} & \textbf{HE-friendly} \\
                 \hline
                %RN50 & \multirow{3}{*}{CIFAR-10} & 99.58 & 99.50 \\
                %RN101 & & 99.80 & 99.58 \\
                %CNT   & & 99.73 & 82.10 \\
                %\hline
                RN50 & \multirow{1}{*}{Places365} &  55.00 & 54.60 \\
                \hline
                RN50 & \multirow{3}{*}{ImageNet} & 80.86 & 76.20 \\
                RN101 & & 81.88 & 77.00 \\
                CNXT & & 82.10 & 79.09 \\
                \hline
        \end{tabular}
        \end{table}
    \end{minipage}
\end{table}

\begin{table}[t!]
    \centering
    \vspace{-20pt}
    \begin{minipage}[t]{0.47\textwidth}
        \begin{table}[H]
            \footnotesize
            \centering
            \caption{The effect of eliminating \glspl{SC} on latency and $\#$bootstraps for HE-friendly ResNet-50 with fixed degree poly-activations.}
            \label{tab:skiptiming}  
            \begin{tabular}{|c|c|c|c|c|}
                \hline
                 \textbf{Deg.} & \multicolumn{3}{c|}{\textbf{\#Bootstraps}} & \textbf{Total} \\
                 & \textbf{w/ skip}  & \textbf{w/o skip} & \textbf{ratio} & \textbf{speedup} \\
                \hline
               2          & $3{,}328$ & $896$       & 3.71 & 2.53 \\ 
            %   4          & $1{,}376$ & $1{,}088$ & 1.26 & \todo{1.8} \\ 
               8          & $5{,}136$ & $3{,}968$ & 1.29 & 1.37  \\ 
               16         & $8{,}480$ & $6{,}976$ & 1.21 & 1.23  \\ 
               18         & $8{,}480$ & $6{,}976$ & 1.21 &  1.21 \\ 
            \hline
        \end{tabular}
        %\vspace{10pt}
    \end{table}
    \end{minipage} \hfill
    \begin{minipage}[t]{0.49\textwidth}
       \begin{table}[H]
            \footnotesize
            \centering
            \caption{Performance on large images ($224 \times 224 \times 3$) for ResNet (RN) models with degree $18$ polynomial activations. BTS - Bootstraps}
            \label{tab:models}
            \begin{tabular}{|l|c|c|c|c|}
                \hline
                \textbf{Arch.} & \textbf{CPU}   &  \textbf{GPU} & \textbf{$\#$BTS} & \textbf{GPU Mem} \\
                        & \textbf{(Min)} & \textbf{(Min)} &             & \textbf{(GB)} \\
                \hline
                RN$18$    & 29.48  &  7.43   & $1{,}184$ & 127 \\ 
                RN$50$    &  152.0   &  31.03 &  $8{,}480$ &  173.37\\
        %    RN$50$ NS &  N/A   &       &     &   \\
                RN$101$   & 295.64  & 57.31  &  $13{,}424$ &  142.1  \\
                RN$152$   & 390.65  & 75.81  & $19{,}424$  & 109.7 \\   
            %    RN$50$ AS &  N/A   &       &     &   \\ \hline
             %   RN$50_{32 \times 32}$ &          &       &     &    \\
                \hline
        \end{tabular}
        \end{table}
    \end{minipage}
\end{table}

\paragraph{A Comparison with the SOTA.}
Lee et al. \cite{pmlr-v162-lee22e} were the first to report promising latency results when considering secure evaluation over ResNet-20/110. For example, it takes only 37 minutes to run ResNet-20 on a single CPU thread. However, this implementation is tailored to datasets such as CIFAR-10/100 with small images of size $32 \times 32$. %This allowed Lee \cite{pmlr-v162-lee22e} to fit most computations into a single ciphertext and reduce the overall number of bootstraps and other operations. 
%Nevertheless, We use a different objective function, by training a low-degree large polynomial CNNs, which allows us to perform results of secure prediction over large images ($224 \times 224$).
%While our reported latency is higher than theirswall-clock measurements is longer than XXX, inference over large images is, as expected, much more challenging. 
%We evaluated computations via HElayers SDK since it relies on a data structure called tile-tensors \cite{helayers}, which enables efficient running of large networks over large images spread across several ciphertexts.
Our approach relies on a different objective function, by training low-degree large polynomial CNNs, which allows us to perform secure prediction over large images.
By using the HElayers SDK we can support larger images that do not fit within a single ciphertext and provide a more generic solution.
%We use a different objective function, training a large CNN over a large-scale dataset with large images ($224 \times 224 \times 3$). The orthogonal question of how to perform this task efficiently in HE is not discussed in this paper. To this end, we use HElayers, which has easy-to-use APIs. Moreover, it relies on a data structure called tile-tensors \cite{helayers}, which enables efficient running of large networks over large images spread across several ciphertexts. One disadvantage of HElayers is that it is not tailored to small images \cite{pmlr-v162-lee22e}. Its developers say they plan to adjust HElayers to these use-cases in the future.

%% Temp for submit draft
% \section{{\color{red}The Potential} of HE-friendly Foundation Models for Transfer
\section{The Potential of HE-friendly Foundation Models for Transfer
Learning\label{sec:app}} % {HE-friendly \acrlong{ZSL}}

Our method enables the training of large-scale polynomial CNNs on unencrypted data, which can then be leveraged for new two capabilities in secure transfer learning: %
(i) \textbf{\gls{ZSL} as an alternative for training over encrypted data:} direct training of polynomial models under HE poses two main challenges: Firstly, certain training techniques such as batch normalization, gradient clipping, and CE loss, which are non-polynomial, are not natively supported under HE. Secondly, the solution's latency increases linearly with the number of training iterations. Hence, inspired by recent advancements in foundation models \cite{Foundationopportunities}, we propose training an \textbf{HE-friendly foundation model} on large-scale unencrypted data. This model can be applied to unseen encrypted data for downstream tasks without additional training, which allows for avoiding the limitations of polynomial training. %
(ii) \textbf{Polynomial pre-trained models for transfer learning:} Previous studies have utilized frozen pre-trained non-polynomial models as feature extractors, followed by secure training of logistic regression on top of these representations \cite{lee2022}. This technique has two major drawbacks: First, the non-polynomial pre-trained model can not be encrypted via HE and can not be applied over encrypted data at inference, as it is not polynomial. Our method from Sect. \ref{sec:method} opens the door to solve this problem by employing polynomial (HE-friendly) pre-trained models. %

Second, when considering fine-tuning over encrypted data, instead of utilizing pre-trained models as \textbf{freeze} feature extractors, our technique allows E2E secure fine-tuning. This approach allows \textbf{optimizing the weights} of the pre-trained polynomial model.% itself, leading to potentially enhanced performance in certain scenarios \ref{bitfit}.

%%%%%%%%%%%%%%%%%%%%%%%%%%%%%%%%%%%%%%%%%%%%%%
%%%%%%%%%%%%%%%%%%%%%%%%%%NEW done%%%%%%%%%%%%%%%%%%%%%%%%%%%%%%%%
%%%%%%%%%%%%%%%%%%%%%%%%%%%%%%%%%%%%%%%%%%%%%%%%%%%%%%%%%%%%%%%%%

\paragraph{Experimental Results - Polynomial CLIP.}
We focus on a specific model that uses contrastive learning~\cite{cr1,cr2,cr3} -- CLIP \cite{CLIP2021}. 
Fig. \ref{fig:CLIP} illustrates the training and inference flows of using an HE-friendly CLIP model. A \gls{SP} starts from a pre-trained CLIP model, modifies its visual encoder to become polynomial, and trains the network. To use the trained model, the service provider shares the text encoder with the clients and uses the visual encoder locally. We consider our polynomial visual encoder as the \textbf{first polynomial foundation model}. Despite its relatively small size (23M parameters), it is one of the largest polynomials trained, and it was established on 400M (image, text) pairs, and adapted to become polynomial encoder through the ImageNet-1K dataset. % ($\sim{}1$M images).

We chose RN50-CLIP due to its image encoder, which is based on a variation of ResNet-50 that we have managed to transform into an HE-friendly model. 
As in previous studies in HE, which incorporated the softmax computation with the client side, we also calculated the output attention-pool layer and cosine similarity on the client side.

Due to the lack of the massive training data used by CLIP, we adapted the original model into an \gls{HE}-friendly model by fine-tuning the model through ImageNet, with the prompts provided by the authors of CLIP. We then evaluated the model on four datasets, that were used by \cite{CLIP2021}: CIFAR-10, CIFAR-100, OxfordPet  and STL10. 
%All of the images are used in a resolution of $224 \times 224 \times 3$.
The results are shown in Tab. \ref{tab:zeroshot}, where we see that even though the \gls{HE}-friendly adaptation process has been managed by a low-resource training set, its prediction capabilities on unseen data remain comparable in various tasks. Moreover, some degradation is expected as the authors of CLIP have noted that using pre-trained models trained on ImageNet achieve lower transfer scores compared to CLIP-based models (see \cite{CLIP2021}, Fig. 12). This is a first step towards an \gls{HE} zero-shot transfer and even few-shot learning on encrypted images. %
% Temp for submitting draft:
Furthermore, similar to CLIP, linear-probing on top of our polynomial model leads to improved accuracy, as observed in the second part of Tab. \ref{tab:zeroshot}. In contrast to previous adaptations of pre-trained models in HE, our approach enables complete encryption of the entire model, as mentioned in the preceding paragraph.
% Furthermore, similar to CLIP, linear-probing on top of our polynomial model can lead to improved results, as observed in the second part of Tab. \ref{tab:zeroshot}. In contrast to previous adaptations of pre-trained models in HE, our approach enables complete encryption of the entire model, as mentioned in the preceding paragraph.

\begin{table}[t!]
    \centering
    \begin{minipage}[t]{0.44\textwidth}
        \begin{table}[H]
        \centering
        \caption{ZSL and linear probe performance of HE-friendly RN50-CLIP fine-tuned on ImageNet and evaluated on various datasets.}
        {\footnotesize
        \label{tab:zeroshot}
        \begin{tabular}{|@{}l|c|c|c|c|}
    \hline
    \textbf{CLIP}   & \textbf{CIFAR} & \textbf{CIFAR} & \textbf{STL10} & \textbf{Pets} \\
    \textbf{Type}    & \textbf{10}    & \textbf{100}   &  \cite{STL10Dataset} & \cite{OxfordPetsDataset}  \\
    \hline
     \multicolumn{5}{c}{Zero-Shot Classification}\\
    \hline
     RN50     & 75.6  & 41.6  & 94.3  & 85.4 \\
     %CLIP     &       &       &       &      \\
    \cline{1-5}
     \textbf{Poly} & 73.3 & 38.4  & 90.7 & 73.9 \\
    %  \textbf{CLIP} &       &       &       &      \\
    \hline 
      \multicolumn{5}{c}{Linear Probing}\\
    \hline 
      RN50        & 88.7 & 70.3  & 96.6  & 88.2 \\
    %  CLIP   &       &       &       &      \\
    \cline{1-5}
     \textbf{Poly}  & 89.28  & 67.71  & 96.96  & 90.62 \\
    %  \textbf{CLIP}  &        &        &        &      \\
    \hline
    
\end{tabular}
        
    }    
        \end{table}
     \end{minipage}\hfill
    \begin{minipage}[t]{0.51\textwidth}
        \begin{figure}[H]
            \centering
            \includegraphics[width=0.90\linewidth]{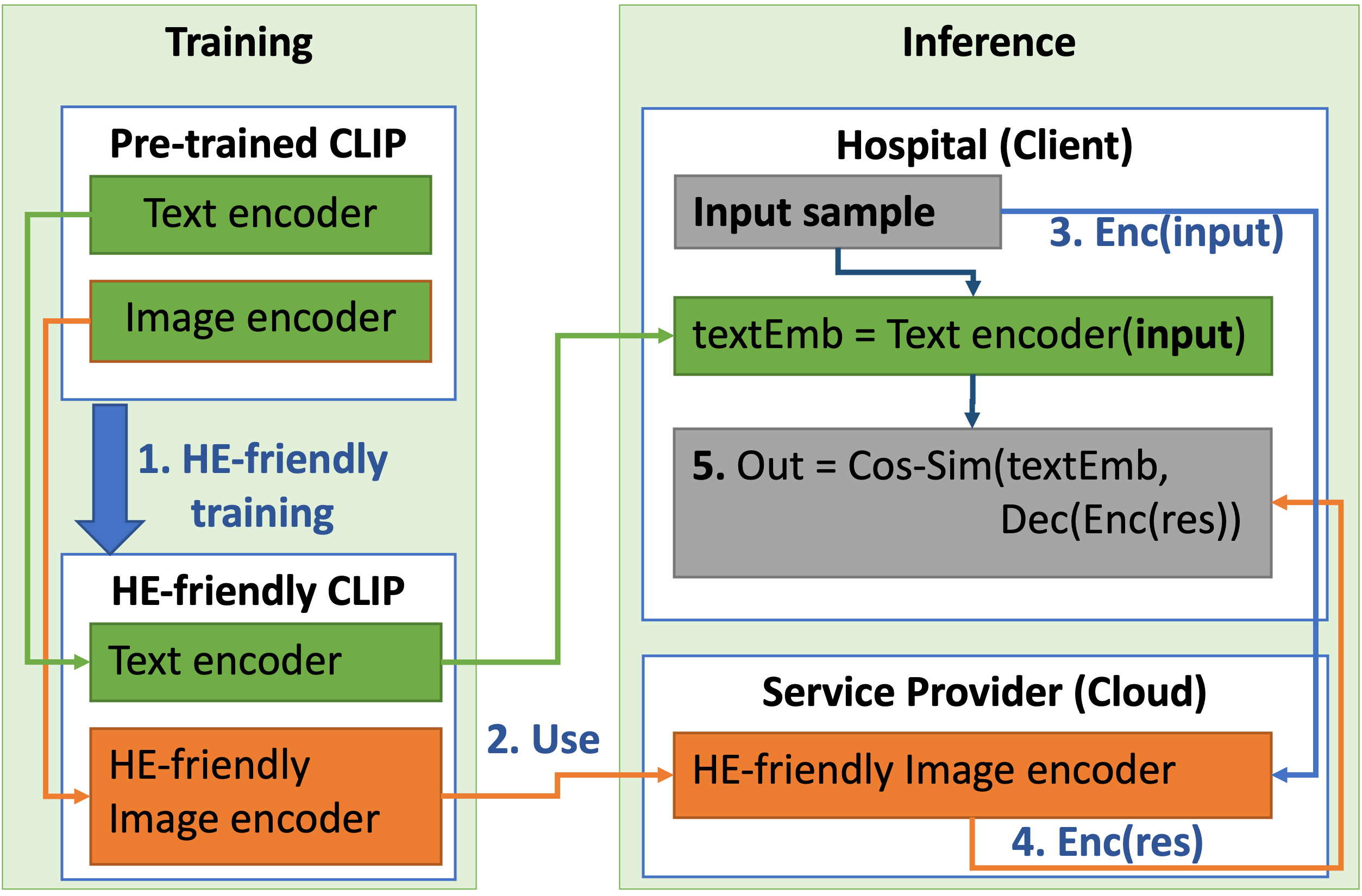}
            \captionof{figure}{HE-friendly CLIP training + secure \gls{ZSL}.}
            \label{fig:CLIP}
        \end{figure}
    \end{minipage}
    \vspace{-25pt}
\end{table}

\vspace{-7pt}

\section{Conclusions}\label{sec:conc}
\vspace{-7pt}
% The question of whether running real-size HE-based polynomial CNNs is \textit{possible} and \textit{practical} has been studied by many researchers over the last decade. However, most studies used some sort of relaxation in the form of client-aided or toy networks and toy datasets to achieve this goal. We answer the above question affirmatively, working on real-size datasets with standard-size images of $224 \times 224$ and modern CNNs such as ResNet and ConvNeXt. We suggest a method for training HE-friendly networks with decent performance and demonstrate that evaluating them under HE is practical. Specifically, we run a ResNet-18/50/101/152 secure prediction in 7, 31, 57, and 75 minutes, respectively, on a GPU with 128-bit security. 

% In addition, we discuss two insights that can further improve the performance of secure predictions, namely, special handling of \glspl{SC} and working with different backbones. For example, we explain the benefits of using ConvNeXt in the context of HE. Finally, our research opens the door for alternatives to the expensive training under HE. This can be achieved through the introduction of ZSL-based methods, or by leveraging new capabilities in transfer learning.

The question of whether running real-size HE-based polynomial CNNs is possible and practical has been studied by many researchers over the last decade. However, most studies used some sort of relaxation in the form of client-aided or toy networks and toy datasets to achieve this goal. We answer the above question affirmatively, working on real-size datasets with standard-size images of $224 \times 224$ and modern CNNs such as ResNet and ConvNeXt. This achievement unlocked thanks to our method to control and minimize the input ranges to the non-polynomial activations. We demonstrate that evaluating them under HE is practical. Specifically, we run ResNet-18/50/101/152 secure prediction in 7, 31, 57, and 75 minutes, respectively, on a GPU with 128-bit security. In addition, we discuss two insights that can further improve secure predictions performance, namely, special handling of SCs and working with different backbones. For example, we explain the benefits of using ConvNeXt in the context of HE. Finally, our research opens the door for alternatives to expensive training under HE. This can be achieved through the introduction of ZSL-based methods, or by leveraging new transfer learning capabilities.

\paragraph{Limitations.}
Our work focuses on training modern polynomial CNNs for HE, but we have not evaluated our method on transformers yet. Transformers face barriers with the Softmax (attention) operation and layer normalization, which are not easily approximated. We plan to address these in future research.

%\begin{ack}
%Use unnumbered first level headings for the acknowledgments. All acknowledgments
%go at the end of the paper before the list of references. Moreover, you are required to declare 
%funding (financial activities supporting the submitted work) and competing interests (related financial activities outside the submitted work). 
%More information about this disclosure can be found at: \url{https://neurips.cc/Conferences/2020/PaperInformation/FundingDisclosure}.

%Do {\bf not} include this section in the anonymized submission, only in the final paper. You can use the \texttt{ack} environment provided in the %style file to autmoatically hide this section in the anonymized submission.
%\end{ack}

%\bibliographystyle{plainnat}
\bibliographystyle{splncs04}
\bibliography{main}
\newpage

\appendix

\section{Homomorphic Encryption}\label{app:he}
We start by describing the high-level background and basic concepts of \gls{HE} schemes.
\gls{HE} schemes allow us to perform operations on encrypted data~\cite{Halevi2017}.
Modern \gls{HE} instantiations such as BGV \cite{bgv}, B/FV \cite{bfv1, bfv2}, and CKKS~\cite{ckks2017} rely on the complexity of the Ring-LWE problem \cite{rlwe} for security and support \gls{SIMD} operations. The \gls{HE} system has an encryption operation $\Enc:\R_1 \rightarrow \R_2$ that encrypts input plaintext from the ring $\R_1(+, *)$ into ciphertexts in the ring $\R_2(\oplus, \odot)$ and an associated decryption operation $\Dec:\R_2 \rightarrow \R_1$. An \gls{HE} scheme is correct if for every valid input $x,y \in \R_1$
\begin{align}
& \Dec(\Enc(x)) = x\\
\label{eq:add}& \Dec(\Enc(x) \oplus \Enc(y)) = x + y \\
\label{eq:mul}& \Dec(\Enc(x) \odot \Enc(y))  = x * y
\end{align}
% {\color{red}
% \begin{align}
% \label{eq:add_and_mul}
% & \Dec(\Enc(x)) = x ,\quad
% & \Dec(\Enc(x) \oplus \Enc(y)) = x + y ,\quad
% & \Dec(\Enc(x) \odot \Enc(y))  = x * y
% \end{align}
% }
and is approximately correct (as in CKKS) if for some small $\epsilon > 0$ that is determined by the key, it follows that $|x - \Dec(\Enc(x))| \le \epsilon$. Eqs.~\ref{eq:add}, and \ref{eq:mul}
are modified in the same way. For this paper, we used CKKS for the experiments.

\section{The HE-based PPML landscape}\label{sec:rel}

The HE-based \gls{PPML} landscape includes interactive/client-aided and non-client-aided solutions. In addition, some solutions involve \gls{NAS} in their design. We briefly review the two terms below.

\paragraph{Client-Aided Solutions. \label{paragraph:client_aided}}
Some \gls{PPML} solutions rely solely on \gls{HE} such as~\cite{helayers, hemet}. They are known as \textit{non-interactive} or \textit{non-client-aided} protocols, while others are known as interactive or client-aided protocols. In the client-aided approach, the server asks the client for assistance with computation, e.g., computing a non-polynomial function such as \ReLU. Here, the server asks the client to decrypt the intermediate ciphertext data, perform the \ReLU computation, and re-encrypt the data using \gls{HE}. This approach is implemented, for example, in GAZELLE~\cite{GAZELLE2018} and nGraph-HE~\cite{HET}. To avoid leakage of intermediate results to the client, the server utilizes a dedicated \gls{MPC} protocol.

The main drawback of client-aided solutions is that the client must stay online during the computation. Moreover, this approach may involve some security risks as detailed in \cite{AkaviaVald21}. \cite{muse} showed that for some cases it can even facilitate the performance of model-extraction attacks. To avoid such attacks, we focus on non-client-aided solutions, where inference computation is performed entirely under encryption, without interaction. 

\begin{table*}[ht!]
    \centering
    \caption[A comparison of \gls{SOTA} \gls{PPML} solutions.]{A comparison of \gls{SOTA} \gls{PPML} solutions. The columns are: \\
    \textbf{Architecture (Arch.):} AlexNet (A); CryptoNets (C); ConvNeXt (CN); DesnseNet-N (DN); 2 hidden layers (H); InceptionNet (I); Industrial (Ind); Lenet-5 (L); MiniONN$^1$ (M); MobileNetV2 (M2); ResNet-N (RN); SqueezeNet (S); VGG16 (V).  \\
    \textbf{Image size (IS):} \priority{0} MNIST ($28 \times 28 \times 1$); 
    \priority{30} CIFAR-10 ($32 \times 32 \times 3$); \priority{50} CIFAR-100 ($32 \times 32 \times  3$);
    \priority{80} Tiny-ImageNet ($64 \times 64 \times  3$);
    \priority{100} ImageNet/CXR$^2$ ($224 \times 224 \times 3$). \\
    \textbf{Non-interactive (NI):} \priority{0} Non-constant round protocol, \priority{50} Constant round protocol, \priority{100} non-interactive protocol. \\
    \textbf{Activation (Act):} \ReLU (R); \ReLU6 (R6), Square (S); Quadratic approximation (Q); quadratic Trainable coefficients (T); Medium-degree approximation (M); High-degree approximation (H); SIgn activation (SI). \\
    \textbf{Limitation (Lim.):} Binarized network (B), Leaks information to the client (L), Not implemented under HE (N), more than two-party (P+), Pruned (smaller) network (Pr). \\
    $^1$ A network from \cite{Liu2017} with 7 convolutional layers, 7 \ReLU layers, 2 mean-Pooling layers, and 1 fully connected layer. \\
    $^2$ Chest X-Ray dataset large images of size $224 \times 224 \times 3$. \\
    $^3$ with approximated Softmax. \\
    $^4$ A reduced SqueezeNet variant. \\
    $^5$ The authors benchmarked ImageNet but did not report accuracy results.
    }
    \label{tab:qual-comparison}
    \begin{tabular}{|l|c|c|c|c|c|c|c|c|}
        \hline
        Solution & NI & Arch. & IS & Sec. bits & Use NAS & Act. & Lim. \\
        \hline
        SecureML \cite{SecureML} & \priority{0} &H & \priority{0} & N/A & \priority{0} & R/S & P+ \\
         
        MP2ML \cite{mp2ml} & \priority{0} & C & \priority{0} & 128 & \priority{0} & R &  \\
         
        MiniONN \cite{Liu2017} & \priority{0} & M & \priority{50} & 128 & \priority{0} & S &  \\
         
        GAZELLE \cite{GAZELLE2018} & \priority{0} & M & \priority{0}/\priority{50} & 128 & \priority{0} & R &  \\
         
        Chameleon \cite{chameleon} & \priority{0} & M & \priority{0}/\priority{30} & 128 & \priority{0} & R & P+ \\
         
        Falcon \cite{falcon} & \priority{0}& M & \priority{0}/\priority{30} & N/A &  \priority{100} & R$^3$ &  \\

        Delphi \cite{delphi} & \priority{0} & M/R32 & \priority{30}/\priority{50} & 128 & \priority{100} & R/R6/Q &  \\

        Cryptflow2 \cite{cryptflow2} & \priority{0} & S/R50/D121 & \priority{100}  & 128 & \priority{0} & R &  \\

         NGraph-HE \cite{HET} & \priority{0} & C/MN & \priority{0}\priority{100} & 128 & \priority{0} & R/T &  \\

        CryptGPU \cite{cryptGPU} & \priority{0} & V/R152 & \priority{0}-\priority{100} & N/A & \priority{0} & R & P+ \\

        Gala \cite{gala} & \priority{0} & A/V/R152 & \priority{0}/\priority{30} & 128 & \priority{0} & R & \\
        
        AriaNN \cite{AriaNN} & \priority{0} & A/V/R18 & \priority{0}-\priority{100} & 128 & \priority{0} & R & \\

        Hunter \cite{hunter} & \priority{0} & A/V/R32 & \priority{0}-\priority{100} & N/A & \priority{0} & R & Pr \\
        
        \cite{large}     & \priority{0} & MN/R50 & \priority{100} & 128 & \priority{0} & R6 & L \\

         \hline
         
        Deepsecure \cite{deepsecure} & \priority{50} & $\sim{}$C & \priority{0} & 128 & \priority{100} & R &  \\

         XONN \cite{xonn} & \priority{50} & V & \priority{0}/\priority{30} & N/A & \priority{0} & R & B \\
         
         \hline

         CryptoNets \cite{CryptoNets2016} & \priority{100} & C & \priority{0} & N/A & \priority{0} & S  & \\
         
         \cite{secure_outsourced} & \priority{100} & $\sim{}$C & \priority{0} & 80 &\priority{0} & S & \\

        RedSEC \cite{redsec} & \priority{100} & A & \priority{0}-\priority{100} & 128 & \priority{0} & SI & B \\

        CHET \cite{chet_compiler} & \priority{100} & L/Ind/S$^4$ & \priority{30}/\priority{50} & 128 & \priority{0} & Q & \\

        HEMET \cite{hemet} & \priority{100} & A/S$^4$/I & \priority{30}/\priority{50} & 128 & \priority{100} & Q &  \\

        HElayers \cite{helayers} & \priority{100} & A/S & \priority{100} & 128  & \priority{0} &  T & \\

        \hline

        \cite{relu1} & \priority{100} & R20 & \priority{30} & 111.6  & \priority{0} &  H & \\
        
        AutoFHE \cite{autoFHE} & \priority{100} & R56 & \priority{30} & 128 & \priority{0} &  H & Pr \\
        
        \cite{lee2021precise} & \priority{100} & V19/R152 & \priority{30}-\priority{100} & N/A & \priority{0} & H & N \\

        Sisyphus \cite{sisyphus} & \priority{100} & R20/R44 & \priority{30}-\priority{80} & N/A & \priority{0} & Q & N \\

        \cite{pmlr-v162-lee22e} & \priority{100} & R20/R110 & \priority{30}-\priority{50} & 128 & \priority{0} & H & \\
        
        HyPHEN \cite{hyphen} & \priority{100} & R20/R44 & \priority{30}-\priority{50}$^5$ & 128 & \priority{0} & H & \\

        \hline

         Ours & \priority{100} & A/V/MN/R32 & \priority{100} & 128 & \priority{0} & M & \\
         \hline
    \end{tabular}
\end{table*}

\paragraph{\gls{NAS}.}
A recent line of work on privacy-preserving \gls{NAS}~\cite{deepreduce, delphi, cryptonas, Autoprivacy} aims to find more efficient \gls{PPML} architectures by reducing the number of non-polynomial primitives. Our training method is orthogonal to this research direction, because a \gls{NAS}-generated network contains non-polynomial elements, which we can address with our methods. In fact, our method can benefit from the reduced number of non-polynomial layers. Another reason for seeing \gls{NAS} as an orthogonal approach is that it assumes the existence of a pre-trained (foundation) model, which spares us a costly training operation required for novel \gls{NAS}-based networks.

Tab. \ref{tab:qual-comparison} presents a rough comparison of \gls{SOTA} HE-based \gls{PPML} inference solutions. Other surveys can be found in \cite{survey1, survey2}. The first group contains interactive/client-aided solutions that are based on a combination of \gls{HE} and \gls{MPC} techniques such as \glspl{GC}, \glspl{OT} or \gls{SS}. The advantage of these solutions is that they can use non-polynomial operations such as \ReLU and enable the running of large models such as ResNet-152, while the model performance stays stable. On the other hand, they often involve high bandwidth, which clients try to avoid. The second group involves a constant number of iterations that is independent of the network architecture, thereby reducing the bandwidth costs. The largest network reported for these solutions is VGG16 over a medium size dataset - CIFAR-10. 

The third and fourth groups include non-interactive, non-client-aided solutions that are based only on \gls{HE}, which is also the focus of this paper. The third group includes solutions without an \gls{HE} operation known as bootstrapping. Here, the largest evaluated networks are AlexNet and SqueezeNet over data with large images of $224 \times 224 \times 3$. To evaluate larger networks, a support for bootstrapping is required. The solutions of the last group are the most relevant for our study. These studies consider \gls{HE} solutions with bootstrapping, allowing them to evaluate large networks such as ResNet-20 - ResNet-152. However, the reported results either did not have 128-bits security \cite{relu1, lee2021precise} as recommended by the standard \cite{standard}, or did not implement their solution using \gls{HE} \cite{lee2021precise}. In the following paragraphs, we show that there is a big gap between a cleartext model and an encrytpted model. Finally, we compare our work to the work of \cite{pmlr-v162-lee22e}, who reported good performance and latency of a ResNet-50 solution over CIFAR-10. In contrast, we show an unprecedented scaled version of ResNet-50 over the large images of ImageNet along with some other applications of our solution in the encrypted domain. We provide an additional comparison to \cite{pmlr-v162-lee22e} in Sect. \ref{sec:exp}.

The above solutions focus mainly on improving latency and bandwidth while maintaining a decent ML performance. In contrast, other works, such as this paper and \cite{baruch2021fighting} focus mainly on generating HE-friendly models.

\section{HE-based experiments setup}\label{sec:hesetup}

\paragraph{Performance of Predictions over Encrypted Data.}
For the following experiments, the models running time is reported for both the GPU and the CPU hardware, which we describe in App. \ref{sec:hesetup}. The GPU runs were based on an average of $500$ samples per dataset, while the CPU runs were based on only $10$ samples due to the longer processing time. In addition, we evaluate the accuracy of the model when applied to encrypted and unencrypted data, and reach an MSE in the range $[1e^{-12}, 1e^{-10}]$. We evaluated our models under HE using the HElayers SDK. Tab. \ref{tab:models} summarizes the latency and memory results of ResNet-18, 50, 101 and 152 with 128-bit security.

For the experiments, we evaluated two different setups: one using a CPU-based system with 32 CPU cores, 32 threads, and 200 GB of memory that were allocated to the tested process. In this setup, either AMD EPYC 7763 64-Core processor or an Intel Xeon E5-2667v2 processor were used. The second setup included both CPU and GPU resources, where the CPU specification was similar to the first setup, but an additional single NVIDIA A100-SXM4-80GB GPU with 80GB of memory was used in some parts of the computation.
In these environments, we run HElayers version $1.52$ and set the underlying HE library to HEaaN. The concrete HE parameters were set as follows: We used ciphertexts with $2^{15}$ coefficients, a multiplication depth of $12$, fractional part precision of $42$, and integer part precision of $18$. This context allows us to use up to $9$ multiplications before bootstrapping is required. The security parameters were set to provide a solution with $128$-bits security.

When we report results for running the models on GPU, we actually mean that we are utilizing a combination of both CPU and GPU. Specifically, the bootstrapping and polynomial activation calculations are performed on the GPU, while the remaining layers of the models are run on the CPU. This is because the polynomial HE-friendly models have high memory requirements that cannot be met by running solely on GPU.

% \begin{figure}[t!]
%     \centering
%     \includegraphics[width=0.80\linewidth]{figures/ranges_per_activation.png}
%     \caption{Input ranges degradation per activation layer per epoch, when training with range-awareness loss.}
%     \label{fig:ranges-per-training1}
% \end{figure}

% \begin{figure}[t!]
%     \centering
%     \includegraphics[width=0.8\linewidth]{figures/global_ranges.png}
%     \caption{Minimal and maximal ranges per iteration over all layers for train and test data}
%     \label{fig:ranges-per-training2}
% \end{figure}

\section{Extra information}
\label{sec:profiling}

\begin{figure}[b!]
    \centering
    \includegraphics[width=0.80\textwidth]{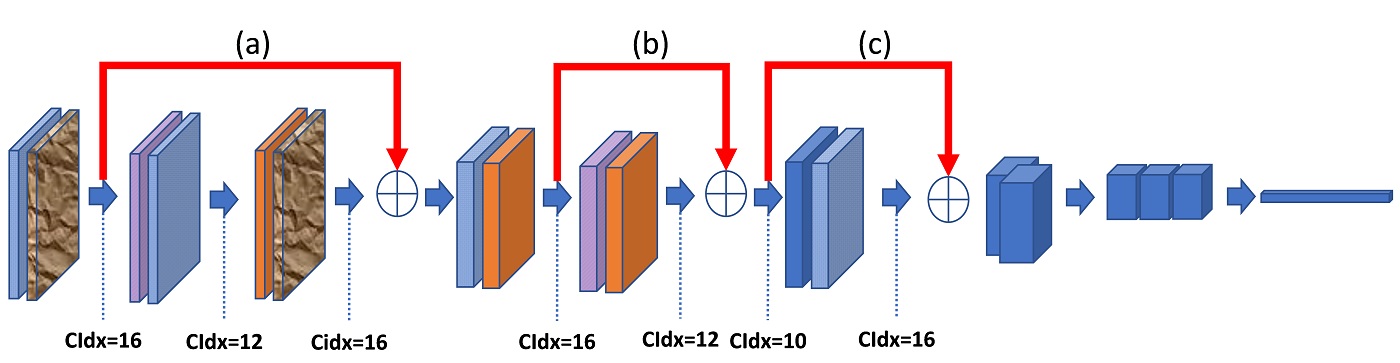}
    \caption{A generic CNN under HE. Every block is a convolutional layer, where the different colors and textures represent different tile tensor shapes \cite{helayers}, i.e., different placement of data inside ciphertexts. The chained index \ChainIdx is indicated below every layer output. The figure includes 3 \glspl{SC} with different latency costs, see text for more details.}
    \label{fig:skip}
\end{figure}

In this section, we provide some graphs that can shed more light on the phenomena discussed in the paper. Fig. \ref{fig:skip} provides a latency breakdown, in percentage, of the layers that mostly contributed to the latency of evaluating ResNet-50 over ImageNet under HE. Specifically, we see that the polynomial activations even though we set their degree to be 18, which is considered small, still consume 43\% of the time. As mentioned above, one research direction to reduce the degrees of the polynomials is to further reduce the layers' input ranges. When considering the potential improvement of reducing SCs, we need to consider the costs of the bootstrapping operations together with the costs of the function g(x), which the SC HE implementation uses. Here, the overall cost is 26\% + 11\% = 37\%.

% We use Figs. \ref{fig:ranges-per-training1} and \ref{fig:ranges-per-training2} to expand some of the graphs from Fig. \ref{fig:compareActivations}. Specifically, Fig. \ref{fig:ranges-per-training1} qualitatively demonstrates what happens to the input ranges of the different layers at every epoch of the training process (Alg. \ref{alg:method}, Step 2), when training with range-awareness loss. Fig. \ref{fig:ranges-per-training2} shows the minimal and maximal 
% values per iteration over all layers for train and test data. In this experiment, we see that the test set bounds are smaller than the train set bounds. Still, Alg. \ref{alg:method} Step 4 tries to ensure that this will always, practically speaking, be the case by estimating the statistically confidential interval for the range of the approximated polynomials' input. 

%Finally, 
We use Fig. \ref{fig:skip} illustrates the logic behind Alg. \ref{alg:scplacement}. It shows a generic \gls{CNN} implemented under \gls{HE}. Here, every block is a convolutional layer, where the different colors and textures represent different tile tensor shapes \cite{helayers}, i.e., different placement of data inside ciphertexts. Below the output of every layer, we indicate the chain index \ChainIdx of the output ciphertexts. The figure includes 3 \glspl{SC}, where the cost of \gls{SC} (a) is low since the chain indices of the two operands match $\ChainIdx(x) = \ChainIdx(f(x))$ and also the tile tensors shapes match; (b) has a higher cost because the chain indices do not match, and because $\ChainIdx(x) > \ChainIdx(f(x))$ a bootstrap operation is required; Finally, the cost of (c) is also high because both the tile tensor shapes and the chain index do not match. However, here, $\ChainIdx(g(x)) < \ChainIdx(f(x))$ so only a HE \ReScale is required and not a \Bootstrap operation.

% \section{Polynomial Evaluation under HE.}
% Evaluating polynomial approximations under HE is not a trivial matter. High-degree polynomials often suffer from extremely small coefficients, below the precision defined by the HE scheme. For example, consider a degree-20 polynomial $p(x)$ that approximates ReLU over inputs in the range $[-10,10]$, and consider its highest monomial $coef \cdot x^{20}$, where $coef$ is the leading coefficient. For an input of $x=10$, we get that $x^{20}=10^{20}$, where the expected result (for the entire polynomial computation) is $ReLU(10)=10$. This forces $coef$ to be very small - in the order of $10^{-20}$. Special handling is required for dealing with such low coefficients and high integers under HE. We address this issue by recursively computing every monomial from its square root, i.e., computing $\sqrt{coef} x^{20/2}$, and squaring the results, instead of computing the input power separately from $coef$. For example, a coefficient to the $x^{16}$ could be $-2.122e-11$.

\section{Polynomial Approximation Per Range}\label{sec:polapporx}

As mentioned in the text, algorithms like Remez \cite{Remez} can derive the optimal \textit{minimax} polynomial $p(x)$ of a fixed degree that has the least error distance from a function $f(x)$ according to some distance metric $d$, i.e., the solution to $\argmin\limits_p{\max\limits_{x \in [a,b]}{d(p(x), f(x))}}$.
Two parameters affect the accuracy of the approximation - the range $[a,b]$ and the polynomial degree. Higher degrees or smaller ranges result in a more accurate approximation. However, a higher degree polynomial might harm the efficiency of the evaluation as larger polynomials require more computations and often increase the overall noise. Therefore, one of our objectives is to reduce the input ranges, which in effect allows us to reduce the polynomial degrees and achieve better efficiency.
%Fig. \ref{fig:compareActivations} Panel (a) compares the polynomial approximation error for \ReLU and \GeLU for different ranges and different polynomial degrees.

\begin{observation}\label{obs:range}
Denote the maximal error of the minimax polynomial of degree $m$ over the range $[a,b]$ by $e_{a,b,m} = \min\limits_p{\max\limits_{x \in [a,b]}{d(p(x), f(x))}}.$ 
Then for all $a_1 < a_2 < b_2 < b_1$ it follows that $e_{a_1,b_1,m} \ge e_{a_2,b_2,m}$.
\end{observation}

\begin{proof}
Let $P_1(x)$ and $P_2(x)$ be the unique minimax polynomials associated with the ranges $[a_1,b_1]$ and $[a_2,b_2]$, respectively, and assume that the observation is false, i.e., that $e_{a_1,b_1,m} < e_{a_2,b_2,m}$, then $P_1$ is also the minimax polynomial for the range $[a_2,b_2]$, which is a contradiction to the fact that $P_2$ is a minimax polynomial.
\end{proof}

\end{document}

%% file: macros.tex
\usepackage[nonatbib,preprint]{neurips_2023}

\usepackage[utf8]{inputenc} % allow utf-8 input
\usepackage[T1]{fontenc}    % use 8-bit T1 fonts
\usepackage{hyperref}       % hyperlinks
\usepackage{url}            % simple URL typesetting
\usepackage{booktabs}       % professional-quality tables
\usepackage{amsfonts}       % blackboard math symbols
\usepackage{nicefrac}       % compact symbols for 1/2, etc.
\usepackage{microtype}      % microtypography

\usepackage{microtype}
\usepackage{graphicx}
\usepackage{booktabs} % for professional tables
\usepackage[utf8]{inputenc} % allow utf-8 input
\usepackage[T1]{fontenc}    % use 8-bit T1 fonts
\usepackage{xurl}           % Allowing line breaks in urls
\usepackage{hyperref}       % hyperlinks
\usepackage[hyperpageref]{backref}

\usepackage{url}            % simple URL typesetting
\usepackage{booktabs}       % professional-quality tables
\usepackage{amsfonts}       % blackboard math symbols
\usepackage{nicefrac}       % compact symbols for 1/2, etc.
\usepackage{subcaption}
\usepackage[english]{babel}
\usepackage{algorithm}

\usepackage{algorithmic}

\usepackage[table,xcdraw]{xcolor}
\usepackage{soul}
\sethlcolor{yellow}
\usepackage{multirow}
\usepackage{mathtools}
\usepackage{amsthm}
\usepackage{xspace}
\usepackage{enumitem}

\usepackage{doi}
\usepackage[acronym]{glossaries}

\glsdisablehyper

\usepackage{siunitx}
\usepackage{tikz}
\newcommand*{\priority}[1]{\begin{tikzpicture}[scale=0.09]%
    \draw (0,0) circle (1);
    \fill[fill opacity=0.99,fill=black] (0,0) -- (90:1) arc (90:90-#1*3.6:1) -- cycle;
    \end{tikzpicture}}
\theoremstyle{plain}
\newtheorem{theorem}{Theorem}[section]

\theoremstyle{definition}

\theoremstyle{remark}

\newtheoremstyle{observation}% name of the style
  {}% space above
  {}% space below
  {\itshape}% body font
  {}% indent amount
  {\bfseries}% theorem head font
  {.}% punctuation after theorem head
  {.5em}% space after theorem head
  {}% theorem head spec (can be left empty, meaning `normal')

\theoremstyle{observation}% apply the new style
\newtheorem{observation}[theorem]{Observation}

\newcommand{\comment}[1][]{\textcolor{blue}}

\newcommand{\todo}[1][]{\textcolor{red}}

\usepackage{lineno}
\usepackage{hyperref}

\hypersetup{
    colorlinks,
    linkcolor={black!50!black},
    citecolor={black!50!black},
    urlcolor={black!80!black}
}

\usepackage{academicons}
\newcommand{\orcid}[1]{\href{https://orcid.org/#1}{\textcolor[HTML]{A6CE39}{\aiOrcid}}}

\DeclareMathOperator*{\argmin}{arg\,min}

\newcommand{\Math}[1]{\ensuremath{#1}\xspace}

\newcommand{\ReLU}{\Math{\operatorname{ReLU}}}
\newcommand{\GeLU}{\Math{\operatorname{GELU}}}
\newcommand{\Enc}{\Math{\operatorname{Enc}}}
\newcommand{\Dec}{\Math{\operatorname{Dec}}}
\newcommand{\ChainIdx}{\Math{\operatorname{CIdx}}}
\newcommand{\ReScale}{\Math{\operatorname{ReScale}}}
\newcommand{\Bootstrap}{\Math{\operatorname{Bootstrap}}}
\newcommand{\R}{\Math{\mathcal{R}}}
\newcommand{\NPL}{\Math{\operatorname{NPL}}}

\newcommand{\regterm}{\Math{rl}}
\newcommand{\NPLSize}{\Math{L}}
\newcommand{\M}{\Math{\mathcal{M}}}
\newcommand{\trainset}{\Math{\mathcal{DS}_{train}}}
\newcommand{\rangeset}{\Math{\mathcal{DS}_{trainRange}}}

\newcommand{\A}{\Math{\mathcal{A}}}
\newcommand{\analyzer}{\Math{\operatorname{HEAnalyzer}}}
\newcommand{\costsMatrix}{\Math{costsMatrix}}

\newacronym{CE}{CE}{Cross-Entropy}
\newacronym{CXR}{CXR}{chest X-Ray}
\newacronym{DL}{DL}{Deep Learning}
\newacronym{CNN}{CNN}{Convolutional Neural Network}
\newacronym{GC}{GC}{Garbled Circuits}
\newacronym{HE}{HE}{Homomorphic Encryption}
\newacronym{FHE}{FHE}{Fully Homomorphic Encryption}
\newacronym{KD}{KD}{Knowledge Distillation}
\newacronym{ML}{ML}{Machine Learning}
\newacronym{MPC}{MPC}{Multi-Party Computation}
\newacronym{NAS}{NAS}{Network Architecture Search}
\newacronym{SC}{SC}{Skip-Connection}
\newacronym{OT}{OT}{Oblivious Transfer}
\newacronym{PPML}{PPML}{Privacy-Preserving Machine Learning}
\newacronym{ResNet}{ResNet}{Residual Net}
\newacronym{SIMD}{SIMD}{Single Instruction Multiple Data}
\newacronym{SS}{SS}{Shared Secret}
\newacronym{ZSL}{ZSL}{Zero-Shot Learning}
\newacronym{SOTA}{SOTA}{State-of-the-Art}
\newacronym{SP}{SP}{Service Provider}
\newacronym{FFN}{FFN}{Feed-Forward Network}